\documentclass[journal,table,dvipsnames]{IEEEtran}

\usepackage[hyphens]{url}
\usepackage{amsfonts}

\usepackage{graphicx}
\usepackage{xcolor} 
\usepackage{algorithm}
\usepackage{algpseudocode}
\usepackage{listings}
\usepackage{wrapfig}
\usepackage{booktabs}
\usepackage{verbatim}
\usepackage{subcaption}
\usepackage{cleveref}
\usepackage{microtype}

\definecolor{codegreen}{rgb}{0,0.6,0}
\definecolor{codegray}{rgb}{0.5,0.5,0.5}
\definecolor{codepurple}{rgb}{0.58,0,0.82}
\definecolor{backcolour}{rgb}{0.95,0.95,0.92}

\lstdefinestyle{mystyle}{
    backgroundcolor=\color{backcolour},   
    commentstyle=\color{codegreen},
    keywordstyle=\color{magenta},
    numberstyle=\tiny\color{codegray},
    stringstyle=\color{codepurple},
    basicstyle=\ttfamily\footnotesize,
    breakatwhitespace=false,         
    breaklines=true,                 
    captionpos=b,                    
    keepspaces=true,                 
    numbers=left,                    
    numbersep=5pt,                  
    showspaces=false,                
    showstringspaces=false,
    showtabs=false,                  
    tabsize=2
}

\begin{document}

\title{From Code to Play: Benchmarking Program Search for Games Using Large Language Models}

\author{Manuel Eberhardinger,
        James Goodman,
        Alexander Dockhorn, 
        Diego Perez-Liebana,
        Raluca D. Gaina,
        Duygu Çakmak, 
        Setareh Maghsudi,
        Simon Lucas
        
        \thanks{Manuel Eberhardinger is with the Institute of Applied AI, Stuttgart Media University, Nobelstr. 10, 70569 Stuttgart, Germany \textit{(Corresponding Author; email: eberhardinger@hdm-stuttgart.de)}}
        \thanks{James Goodman, Diego Perez-Liebana, Raluca D. Gaina and Simon Lucas are with the School of Electronic Engineering and Computer Science, Queen Mary University of London, E1 4NS London, U.K.}
        \thanks{Alexander Dockhorn is with the Institute for Information Processing, Leibniz University Hannover, Appelstr. 9A, 30167 Hannover, Germany }
        \thanks{Duygu Çakmak is with Creative Assembly, RH12 1JW Horsham, U.K. }
        \thanks{Setareh Maghsudi is with the Chair of Learning Technical Systems, Ruhr-University Bochum, Universitätsstr. 150, 44801 Bochum, Germany}
        }

\maketitle
\IEEEpeerreviewmaketitle

\begin{abstract}
Large language models (LLMs) have shown impressive capabilities in generating program code, opening exciting opportunities for applying program synthesis to games. In this work, we explore the potential of LLMs to write usable code for a wide range of gaming applications, focusing on two programming languages, Python and Java. We use a hill-climbing algorithm, where the mutations and seeds of the initial programs are controlled by LLMs.
For Python, the framework covers various game-related tasks, including five miniature versions of Atari games, ten levels of \emph{Baba is You}, an environment inspired by \emph{Asteroids}, and a maze generation task. For Java, the framework contains 12 games from the TAG tabletop games framework. Across 29 tasks, we evaluated 11 language models for generating Python and Java code. 
Our findings suggest that the performance of LLMs depends more on the task than on model size. In addition, the experiments show that running the program search algorithm multiple times with fewer iterations is better than a single run with more iterations.

\end{abstract}

\begin{figure}[tb]
    \centering
    \includegraphics[width=\linewidth]{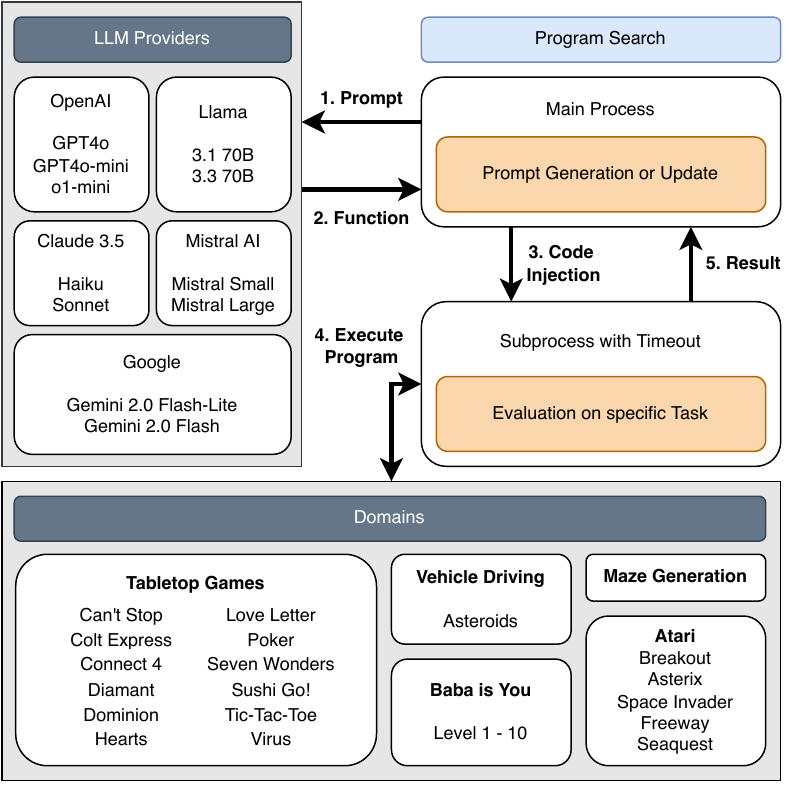}
    \caption{The general framework for program search begins by generating an initial task prompt (1), which is processed by one of the integrated LLMs to produce a function (2). This function is then evaluated within a subprocess (3), which executes the program in the given task (4) and then the results are reported back to the main process (5). The main process updates the prompt based on the evaluation outcomes and either returns it to the LLM (repeat from 1) for further refinement or concludes if the evaluation criteria are reached. }
    \label{fig:framework-overview}
\end{figure}

\section{Introduction}
Before the emergence of large language models (LLMs) for code \cite{codex}, program synthesis in imperative or object-oriented languages like Python or Java was considered a highly challenging task due to the combinatorial explosion of the search space \cite{program-synthesis}. 
Therefore, most solvable tasks were restricted to simple problem domains such as string manipulation or list sorting, typically implemented within a predefined domain-specific language (DSL) \cite{flash-meta}. Similarly, program synthesis for games was limited to simple problems with well-defined search spaces, achievable only by incorporating high-level game-specific concepts into the DSL \cite{megaman, puzzle-games, lpp, microrts}.

The goal of program synthesis is the automatic generation of programs that satisfy certain requirements, such as input-output examples (e.g. five pairs of unsorted lists and their sorted counterparts), natural language descriptions of the program to be generated or formal specifications such as logical constraints \cite{program-synthesis}.
Automatic construction of programs from high-level programming languages in game research has hardly been explored. Most discussions merely outlined its potential applications \cite{game-dev-program-synthesis}, or focused on the missing aspects of automated game development systems to move from game description languages to programming languages \cite{automated-game-design}.

Recently, methods for LLM-based program search have been introduced for the automatic design of playable games based on program code \cite{gavel, llm-vgdl, llm-unity} and to generate game content through JSON representations \cite{llm-json}. LLMs have also been adapted to synthesize programmatic policies in Python, which are then converted into a DSL suitable for the target environment \cite{llm-guided-pp}, as well as to construct world models in Python that approximate reward and state transition functions for simple games, enabling action plan generation \cite{worldcoder}. 

In this work, we explore the potential of LLM-based program search for a wider range of games without depending on a predefined JSON converter \cite{llm-json} or on predefined specifications such as a DSL (e.g., Ludii \cite{gavel}, the video game description language \cite{llm-vgdl}, or Karel \cite{llm-guided-pp}). Our aim is to enable LLMs to synthesize program code that can be used directly, without requiring additional transformations or prior specifications. We evaluate this approach across different domains using two programming languages: Python and Java. In Python, we focus on synthesizing programmatic agent policies and functions for procedural content generation (PCG). In Java, the method is integrated into TAG, a tabletop games framework, in which LLMs design heuristics for board games \cite{tag}. 

Our goal is not to propose a new method for program synthesis, but to introduce an easy-to-use and extensible framework to evaluate the current performance of LLMs for the synthesis of game-related program code. To achieve this, we have integrated five different LLM providers with a total of 11 different models. For the synthesis of Python code, the framework consists of five miniature versions of Atari games where the input is represented symbolically \cite{minatar}, ten levels of the game \textit{Baba is You}, in which various game mechanics are tested \cite{charity2022keke}, a vehicle driving environment based on the game \textit{Asteroids}, and procedural content generation in the form of mazes. For Java, the framework consists of 12 tabletop games of the TAG framework \cite{tag}. In total, we evaluate the LLMs on 29 different tasks. An overview of our proposed framework, as well as games and LLMs used, is shown in \Cref{fig:framework-overview}.

Our contributions are: 
\begin{itemize}
    \item We perform an empirical study to evaluate the current state-of-the-art of LLM-based program search for games.
    \item We introduce an easy-to-use and extensible framework with 29 tasks that evaluate various aspects of game mechanics. 
    \item We open-source our code upon publication. Currently, only the example prompts for the experiments are available in the repository\footnote{\url{https://github.com/ManuelEberhardinger/Benchmarking-Language-Model-Based-Program-Search-for-Games}}.
\end{itemize}

The structure of the paper is as follows: We review related work on program synthesis for games in \Cref{sec:related-work}. Afterwards, the framework is introduced in \Cref{sec:framework}, followed by a short summary of the LLMs used in \Cref{sec:llms}. \Cref{sec:game-applications} first describes the experimental setup (\Cref{sec:exp-setup}) and then the experiments for the game applications, starting with miniature versions of the Atari games, followed by the Asteriods-inspired vehicle driving experiments. Then we discuss the puzzle game Baba is You is and the maze generation experiments. We conclude the first part of the experiments with an overall evaluation of the generated Python code. The second part of the experiments, starting in \Cref{sec:tag}, focuses on the TAG framework which uses Java as the programming language. 
In addition to the standardised experiments, longer-running experiments that evaluate the impact on the number of iterations for the five games in the Atari domain are reported in \Cref{sec:1000Minatar}. Finally, we discuss practical insights gained from the experiments in \Cref{sec:practical}, and conclude the paper in \Cref{sec:conclusion}.

\section{Related Work}
\label{sec:related-work}
There are a considerable number of studies that use program synthesis approaches for games. Butler et al. used SMT-solvers to search for programs within a Lisp-based DSL, enabling the generation of diverse boss fights in Megaman \cite{megaman} and puzzles for the game Nonograms \cite{puzzle-games}. In \cite{lpp}, a method was introduced for learning combinations of logical programs to solve simple grid-based games like Nim. Cropper et al.  \cite{iggp, iggp2, iggp3} developed a comprehensive benchmark of 50 games to recover game rules from gameplay traces using inductive logic programming (ILP). Furthermore, Evans et al. \cite{sokoban-ilp} applied a differentiable form of ILP to learn interpretable rules for Sokoban. Recently, a method for learning programmatic policies for zero-shot coordination problems in cooperative tasks was introduced and demonstrated in the game Overcooked \cite{gu2024knowpc}. In contrast to learning programmatic policies, there is also work focusing on using program synthesis to explain the decision-making process of game-playing agents \cite{ec-rl}.

Mari\~{n}o et al.~\cite{microrts} utilized program search to develop strategies for the game MicroRTS, comparing the resulting programmatic policies with those created by human developers. Their findings demonstrated that the synthesized programs performed comparably to those written by humans. Subsequent research built on this foundation by introducing improved search techniques, including bilevel search \cite{microrts2}, by guiding the program search \cite{microrts3} or by searching in semantic spaces \cite{semantic-spaces}. Recently, an approach for combining LLMs with local search algorithms was proposed for MicroRTS~\cite{sadmine2024language}, where the authors showed that providing initial programs with LLMs found better solutions faster and improved the performance of the final programs.

In \cite{seven-wonders}, Genetic Programming (GP) is used to search for evaluation functions within a predefined DSL for the board game 7 Wonders. Similarly, Sturtevant and White generate evaluation functions automatically by using reinforcement learning to learn more complex features by combining atomic features for the game Hearts \cite{10.1007/978-3-540-75538-8_11}. This approach resembles our experiments with the TAG framework, where heuristic functions are synthesized; however, we use Java without relying on predefined concepts. GP has also been applied to generate game agents for various scenarios, including a fighting game \cite{fight-gp}, a platformer game \cite{platform-gp}, a puzzle game \cite{OlsWag2023}, and to create explanations for a maze runner agent \cite{gp-maze}. Additionally, Wilson et al.~\cite{cartesian-gp} used Cartesian GP to develop programs for Atari games, processing pixel observations through predefined mathematical, statistical, or list functions.

A recent approach from cognitive science, known as language-informed thinking \cite{world-models}, combines large language models (LLMs) with Bayesian inference. This method enables LLMs to pose questions in natural language, which are then translated into a language of thought \cite{lot} represented as a probabilistic programming language. Grand et al. extended this approach to the board game Battleship, demonstrating that the questions generated by LLMs aligned closely with the performance of human players \cite{battleship}.

Verma et al.~\cite{pirl, pirl2} employed neurosymbolic methods to synthesize programmatic policies for a car racing game, demonstrating that these programs were more robust than neural network policies while achieving comparable rewards. While this approach shares similarities with the vehicle driving experiments in our work, it is more constrained, as the search space is limited to the provided DSL and our vehicle driving problem requires a planning algorithm to be solvable.

In \cite{autumn}, a reactive programming language with a novel program synthesis algorithm is introduced to discover causal structures represented as state machines in combination with program code. They evaluate the proposed method for 2D grid games similar to Super Mario.

Voyager \cite{wang2024voyager} is a lifelong learning agent for the Minecraft environment that uses an LLM to synthesize code in the Mineflayer API\footnote{\url{https://github.com/PrismarineJS/mineflayer/tree/master}}, which is then executable to obtain the actions. In addition, a second LLM is used as a high-level planner to create a task curriculum for the agent. 
Moreover, Ma et al. \cite{eureka} proposed an evolutionary approach using LLMs to synthesize reward functions for complex control tasks, achieving superior performance compared to human-engineered reward functions.

The key distinction between our work and the discussed literature is the use of high-level programming languages (Python and Java), making our approach applicable to a broader range of tasks without relying on predefined building blocks or a programming library. The work that is most similar to ours and is also used for game environments is \cite{worldcoder}, where Python code is synthesized to approximate a world model. However, this work is limited to a single, different type of task.

\section{Framework}
\label{sec:framework}
The general framework we propose is based on a hill-climbing algorithm where the mutations and the seed of the initial program are performed by an LLM \cite{llm-guided-pp, funsearch}. Thus, our framework belongs to the group of neurosymbolic programming methods \cite{chaudhuri2021neurosymbolic}, as we use an LLM to generate programs that are checked for correctness and functionality by symbolic verifiers, in our case the Python interpreter and Java compiler. The overview of the framework is illustrated in \Cref{fig:framework-overview}, which shows the high-level interaction between the different modules and processes. Synthesized program code by LLMs is always executed within a safe subprocess environment, ensuring that the main process can terminate it after a certain time limit to prevent infinite execution of the program.

A detailed description of the complete algorithm is provided in \Cref{alg:framework}. 
Our framework consists of two iterative processes that control the length of the search, defined by the input parameter \verb|iterations|, as well as the number of attempts to repair the program in each iteration defined by the input parameter \verb|maxRepairs|.
Each iteration starts by generating or updating a task prompt to obtain an initial Python or Java function from \Cref{alg:query-llm}. The procedure of \Cref{alg:query-llm} queries the LLM and tries to fix compilation or syntax errors for the given number of \verb|maxRepairs|. The program code is then returned to \Cref{alg:framework} and is injected and executed in a subprocess.
If the function is executed successfully and returns an improved reward, we update the prompt with the achieved evaluation metric and all relevant environment-specific details, such as the action trace of the executed function. If a runtime error occurs, the error description is included in the prompt for refinement of the program, so that it is possible for the LLM to fix these errors in the next iteration.
These steps are repeated iteratively until the evaluation criteria defined by the fitness function are satisfied or the specified number of \verb|iterations| is reached.

We explain the domain-specific adaptations of the framework in the respective chapters in section \ref{sec:game-applications}. While the overall framework is similar for all tasks, domain-specific adaptations are necessary, such as the description of the environment or the game logic, as well as the objective of the game.
\begin{algorithm}
\caption{The algorithm for querying an LLM with the number of attempts to generate/update or repair the program.} \label{alg:query-llm}
\begin{algorithmic}[1]
\Procedure{QueryLlmWithRepair}{prompt, maxRepairs}
\State program $\gets$ \Call{QueryLlm}{prompt}
\State $j \gets 1$
\While{$j <$ maxRepairs \textbf{and not} CanExec(program)}
\State prompt $\gets$ \Call{GetRepairPrompt}{program}
\State program $\gets$ \Call{QueryLlm}{prompt}
\State $j \gets j + 1$
\EndWhile \\
\Return program
\EndProcedure
\end{algorithmic}
\end{algorithm}

\begin{algorithm}
\caption{The algorithm for our proposed framework, which uses the procedure from Algorithm \ref{alg:query-llm} in each iteration to improve or generate a program.}\label{alg:framework}
\begin{algorithmic}[1]
\Procedure{ProgramSearch}{task, iterations, maxRepairs}
\State bestFitness $\gets$ 0
\State $i \gets 1$
\Repeat
    \If{bestProgram}
        \State prompt $\gets$ \Call{UpdateTaskPrompt}{task, bestResult, bestFitness, bestProgram}
    \Else
        \State prompt $\gets$ \Call{GetTaskPrompt}{task}
    \EndIf
    \State program $\gets$ \Call{QueryLlmWithRepair}{prompt, maxRepairs}
    \State result $\gets$ \Call{InjectAndRunCode}{task, program}
    \State fitness $\gets$ \Call{EvaluateFitness}{task, result}
    \If {fitness $>$ bestFitness} 
        \State bestProgram $\gets$ program
        \State bestResult $\gets$ result
        \State bestFitness $\gets$ fitness
    \EndIf
    \State $i \gets i + 1$
\Until{\textbf{not} \Call{CheckCriterion}{fitness} \textbf{and} 
$i <$ iterations}
\\
\Return program
\EndProcedure
\end{algorithmic}
\end{algorithm}

\section{Large Language Models}
\label{sec:llms}
\begin{table}[tb]
    \caption{The used LLMs with the specific version or date of the release.}
    \centering
    \begin{tabular}{lcc}
    \toprule
    LLM & Release Date  \\
    \midrule
    Claude 3.5 Sonnet & 2024-10-22  \\
    Claude 3.5 Haiku & 2024-10-22  \\
    \midrule
    Gemini 2.0 Flash & 2025-02 \\
    Gemini 2.0 Flash-Lite & 2025-02  \\
    \midrule
    Mistral Large & 2024-11 \\
    Mistral Small & 2025-03 \\
    \midrule
    o1-mini & 2024-09-12 \\
    GPT 4o & 2024-08-06  \\
    GPT 4o mini & 2024-07-18  \\
    \midrule
    Llama 3.3 70B & 2024-12-06  \\
    Llama 3.1 70B & 2024-07-23  \\
    \bottomrule
    \end{tabular}
    \label{tab:llm-overview}
\end{table}
For our benchmark, we integrated five LLM providers for generating Python and Java code in our framework, using one smaller and one larger version for each model type. From OpenAI, we utilize models from the GPT-4o family\footnote{\url{https://platform.openai.com/docs/models}}, based on GPT-4 \cite{gpt4}. 
For the new ChatGPT models in the o1 generation, we use o1-mini, which offers performance comparable to o1-preview for coding tasks.\footnote{\url{https://openai.com/index/openai-o1-mini-advancing-cost-efficient-reasoning/}}
We also incorporate the latest models from Mistral\footnote{\url{https://docs.mistral.ai/getting-started/models/models_overview/}}, Claude 3.5\footnote{\url{https://docs.anthropic.com/en/docs/about-claude/models}} from Anthropic, based on Claude 3 \cite{claude-models}, and the 2.0 Gemini Flash models\footnote{\url{https://ai.google.dev/gemini-api/docs/models/gemini}} \cite{team2024gemini}, provided by Google, for both programming languages. Two open source models from the Llama family are used, Llama 3.1 70B \cite{llama} and Llama 3.3 70B, which performs similarly to Llama 3.1 405B for code-related tasks.\footnote{\url{https://www.llama.com/docs/model-cards-and-prompt-formats/llama3_3/}}
Details on the model versions and their release dates are in Table \ref{tab:llm-overview}.


\section{Game Applications}
\label{sec:game-applications}
In the following section, we first describe the experimental setup, followed by the experiments that were conducted for each of our target domains. 

\subsection{Experimental Setup}
\label{sec:exp-setup}
In general, the experiments reported run 10 independent \verb|trials| of \Cref{alg:framework}, and the final recommendation is the best performing agent from these 10 trials. In each trial, 10 iterations of search were performed, with 3 queries used to create/improve or repair the policy.

\subsection{Programmatic Policies: Minatar}
\begin{figure}[t]
\centering
\includegraphics[width=.32\linewidth]{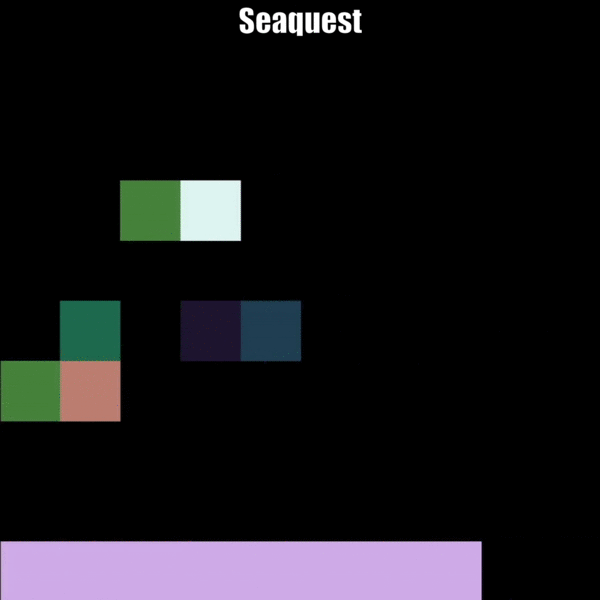} 
\includegraphics[width=.32\linewidth]{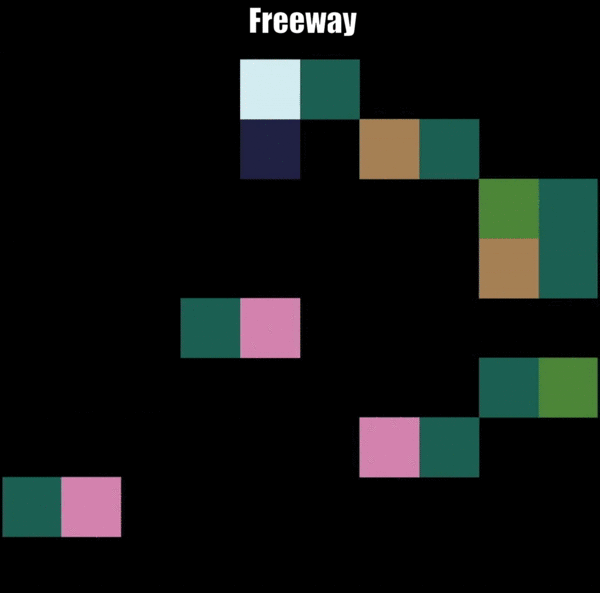}
\includegraphics[width=.32\linewidth]{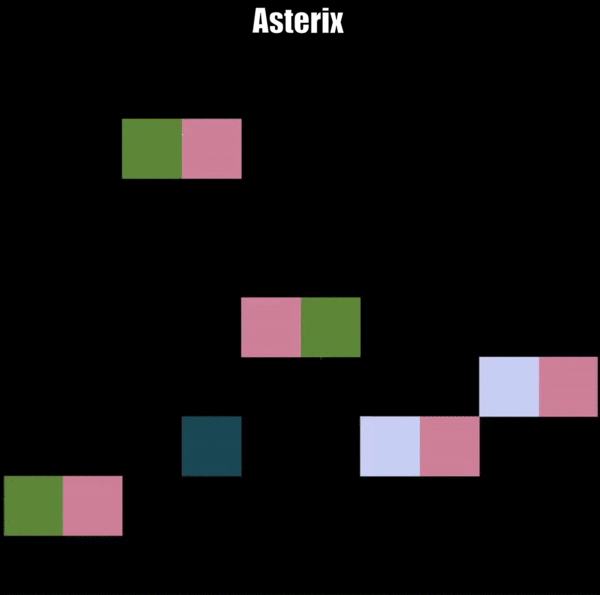}\vspace{0.2cm}
\includegraphics[width=.32\linewidth]{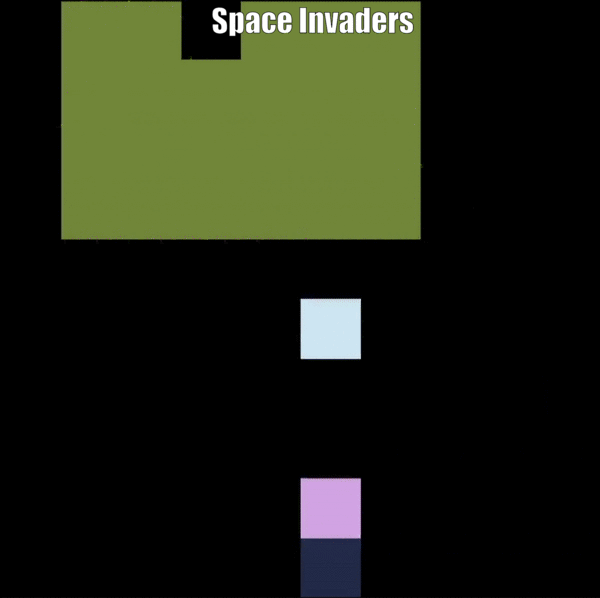}
\includegraphics[width=.32\linewidth]{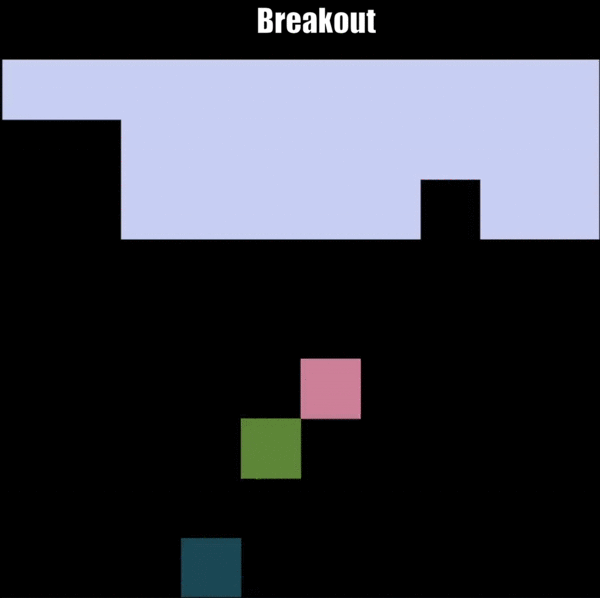}
\caption{The five miniature versions of the Atari games (left to right, top to bottom: Seaquest, Freeway, Asterix, Space Invaders and Breakout), which are used for the synthesis of the programmatic strategies. Each colour represents a different type of object, e.g. the paddle in dark blue, the ball in green and the track of the ball in pink for the game Breakout.}
\label{fig:minatar}
\end{figure}

Minatar \cite{minatar} is a collection of five games that are miniature versions of Atari games. In Minatar, the games are represented as a symbolic state space on a 10$\times$10$\times n$ grid, where $n$ represents the number of channels, and each channel represents an object such as walls, enemies or the agent. Minatar is an ideal test bed for experiments, as the games are more efficient to learn without changing the game mechanics of the original game. Previously, Minatar was used in \cite{ec-rl} to explain the behaviour of agents through program synthesis, but it was only possible to explain short sub-trajectories since enumerative search-based methods were used to search through a predefined domain-specific language that resembles Lisp. In our experiments, we use all available Minatar environments, which are shown in Figure \ref{fig:minatar}. The games are:
\begin{itemize}
    \item \textbf{Seaquest}: The agent controls a submarine and is able to shoot  bullets. The objective is to save as many divers as possible, while also shooting enemy submarines or sharks. Each time an enemy is struck, the reward is increased by one. When the submarine saves the divers, the agent also receives a reward.    
    \item \textbf{Freeway:} The agent controls a chicken that needs to cross a road during rush hour, while avoiding the traffic. For each chicken that crosses the road safely, the agent receives one point.
    \item \textbf{Asterix}: The objective of the game is to collect gold while avoiding enemies. The player gets one point for each collected gold and the game is over when the player is hit by an enemy.    
    \item \textbf{Space Invaders}: The agent controls a cannon and shoots aliens while dodging bullets launched from the alien spaceship. Additionally, the player must prevent the aliens from reaching the bottom of the screen. For each destroyed alien, one point is received.
    \item \textbf{Breakout}: The goal is to destroy all the bricks with the ball by controlling the paddle to bounce the ball off before it goes out off the screen. With each destroyed brick the agent receives one point. 
\end{itemize}

\begin{figure}[tb]
    \centering
    \includegraphics[width=\linewidth]{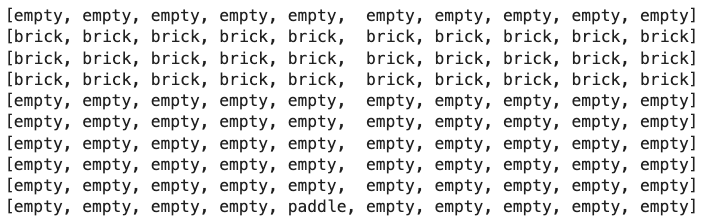}
    \caption{The text description of the state for the Breakout game which is included in the prompt.}
    \label{fig:breakout-state}
\end{figure}

The LLMs were prompted to generate a Python function which can be used as a policy to play the game. The prompt contains information about the game rules, the objective of the agent and also the possible actions of the environment and available game objects. The description of the game was taken from Young and Tian \cite{minatar}. The prompts for the games are available in the code repository\footnote{\url{https://github.com/ManuelEberhardinger/Benchmarking-Language-Model-Based-Program-Search-for-Games}}. The LLM receives only the initial state, which is preprocessed from the state input of the environment, a one-hot encoded 3D array, into a 2D array with text descriptions for each grid cell representing the cell's object. Figure \ref{fig:breakout-state} shows an example of the converted state for Breakout. All other games convert the state in a similar way so that the LLM can process the state input semantically. 

Each of the games tests the LLM for different game concepts. Space Invaders, Breakout and Freeway restrict the agent's movement by only allowing horizontal or vertical movement. Space Invader and Seaquest allow the player to fight the enemy, while in Asterix and Freeway the player can only avoid the enemies. In Asterix, the player must also collect items in order to receive a reward. Seaquest is the most difficult game, as the player has to collect six divers and then reach the surface so that the divers can leave the submarine, but at the same time the player has to shoot down enemies. Breakout is one of the easier games compared to the others, as there are no opponents and the player only has to anticipate where the ball will land in order to bounce it off with the paddle.
\begin{table}[]
    \centering
    \caption{Max Reward of the best program for the Minatar experiments with 50 evaluation episodes.}
    \resizebox{\linewidth}{!}{\begin{tabular}{llccccrr}
    \toprule
    Model & Seaquest & Freeway & Breakout & Asterix & SpaceInvader \\
    \midrule
Claude Sonnet & \cellcolor[rgb]{0.79,0.79,1.00}2.43 & \cellcolor[rgb]{0.64,0.64,1.00}7.22 & \cellcolor[rgb]{0.58,0.58,1.00}18.43 & \cellcolor[rgb]{0.50,0.50,1.00}11.37 & \cellcolor[rgb]{0.50,0.50,1.00}29.07 \\
Claude Haiku & \cellcolor[rgb]{0.68,0.68,1.00}3.68 & \cellcolor[rgb]{0.64,0.64,1.00}7.33 & \cellcolor[rgb]{0.84,0.84,1.00}7.13 & \cellcolor[rgb]{0.70,0.70,1.00}6.78 & \cellcolor[rgb]{0.62,0.62,1.00}21.91 \\
\midrule
Gemini Flash & \cellcolor[rgb]{0.74,0.74,1.00}2.98 & \cellcolor[rgb]{0.60,0.60,1.00}8.07 & \cellcolor[rgb]{0.72,0.72,1.00}12.3 & \cellcolor[rgb]{0.73,0.73,1.00}6.09 & \cellcolor[rgb]{0.55,0.55,1.00}26.34 \\
Gemini Lite & \cellcolor[rgb]{0.95,0.95,1.00}0.58 & \cellcolor[rgb]{0.68,0.68,1.00}6.49 & \cellcolor[rgb]{0.64,0.64,1.00}15.56 & \cellcolor[rgb]{0.78,0.78,1.00}5.06 & \cellcolor[rgb]{0.71,0.71,1.00}16.57 \\
\midrule
Mistral Large & \cellcolor[rgb]{0.89,0.89,1.00}1.3 & \cellcolor[rgb]{0.58,0.58,1.00}8.48 & \cellcolor[rgb]{0.72,0.72,1.00}12.03 & \cellcolor[rgb]{0.72,0.72,1.00}6.48 & \cellcolor[rgb]{0.64,0.64,1.00}20.89 \\
Mistral Small & \cellcolor[rgb]{0.98,0.98,1.00}0.26 & \cellcolor[rgb]{0.59,0.59,1.00}8.2 & \cellcolor[rgb]{0.87,0.87,1.00}5.59 & \cellcolor[rgb]{0.74,0.74,1.00}5.88 & \cellcolor[rgb]{0.61,0.61,1.00}22.41 \\
\midrule
o1 mini & \cellcolor[rgb]{0.50,0.50,1.00}5.73 & \cellcolor[rgb]{0.53,0.53,1.00}9.52 & \cellcolor[rgb]{0.50,0.50,1.00}21.81 & \cellcolor[rgb]{0.54,0.54,1.00}10.39 & \cellcolor[rgb]{0.61,0.61,1.00}22.45 \\
GPT 4o & \cellcolor[rgb]{0.90,0.90,1.00}1.15 & \cellcolor[rgb]{0.50,0.50,1.00}10.05 & \cellcolor[rgb]{0.65,0.65,1.00}15.26 & \cellcolor[rgb]{0.52,0.52,1.00}10.91 & \cellcolor[rgb]{0.59,0.59,1.00}23.83 \\
GPT 4o mini & \cellcolor[rgb]{0.96,0.96,1.00}0.47 & \cellcolor[rgb]{0.58,0.58,1.00}8.44 & \cellcolor[rgb]{0.80,0.80,1.00}8.56 & \cellcolor[rgb]{0.71,0.71,1.00}6.55 & \cellcolor[rgb]{0.53,0.53,1.00}27.45 \\
\midrule
Llama 3.3 70B & \cellcolor[rgb]{0.65,0.65,1.00}4.01 & \cellcolor[rgb]{0.61,0.61,1.00}7.74 & \cellcolor[rgb]{0.80,0.80,1.00}8.85 & \cellcolor[rgb]{0.74,0.74,1.00}5.8 & \cellcolor[rgb]{0.71,0.71,1.00}17.1 \\
Llama 3.1 70B & \cellcolor[rgb]{0.64,0.64,1.00}4.08 & \cellcolor[rgb]{0.60,0.60,1.00}8.02 & \cellcolor[rgb]{0.90,0.90,1.00}4.39 & \cellcolor[rgb]{0.76,0.76,1.00}5.54 & \cellcolor[rgb]{0.78,0.78,1.00}13.04 \\
    \bottomrule
    \end{tabular}}
    \label{tab:minatar-evaluation}
\end{table}

\begin{table}[]
    \centering
    \caption{Average reward of the 10 programs found after each trial. We also report the standard deviation $\sigma$ of the 10 programs after each trial. The min. reward for all games is 0, which indicates that there is a high variance in between the trials.}
    \resizebox{\linewidth}{!}{\begin{tabular}{lccccc}
    \toprule
    Model & Seaquest & Freeway & Breakout & Asterix & SpaceInvader \\
\midrule
Claude Sonnet & \cellcolor[rgb]{0.50,0.50,1.00}1.56$\pm$0.38 & \cellcolor[rgb]{0.79,0.79,1.00}3.32$\pm$2.78 & \cellcolor[rgb]{0.52,0.52,1.00}8.25$\pm$4.1 & \cellcolor[rgb]{0.57,0.57,1.00}6.52$\pm$2.46 & \cellcolor[rgb]{0.50,0.50,1.00}22.38$\pm$5.04 \\
Claude Haiku & \cellcolor[rgb]{0.66,0.66,1.00}1.07$\pm$0.96 & \cellcolor[rgb]{0.76,0.76,1.00}3.8$\pm$3.03 & \cellcolor[rgb]{0.78,0.78,1.00}3.77$\pm$1.85 & \cellcolor[rgb]{0.63,0.63,1.00}5.56$\pm$0.85 & \cellcolor[rgb]{0.68,0.68,1.00}14.5$\pm$5.53 \\
\midrule
Gemini Flash & \cellcolor[rgb]{0.67,0.67,1.00}1.04$\pm$0.75 & \cellcolor[rgb]{0.79,0.79,1.00}3.31$\pm$2.83 & \cellcolor[rgb]{0.66,0.66,1.00}5.82$\pm$3.34 & \cellcolor[rgb]{0.71,0.71,1.00}4.36$\pm$1.22 & \cellcolor[rgb]{0.78,0.78,1.00}9.98$\pm$9.27 \\
Gemini Lite & \cellcolor[rgb]{0.96,0.96,1.00}0.14$\pm$0.19 & \cellcolor[rgb]{0.96,0.96,1.00}0.65$\pm$1.95 & \cellcolor[rgb]{0.60,0.60,1.00}6.8$\pm$5.25 & \cellcolor[rgb]{0.79,0.79,1.00}3.13$\pm$1.69 & \cellcolor[rgb]{0.89,0.89,1.00}5.02$\pm$4.12 \\
\midrule
Mistral Large & \cellcolor[rgb]{0.91,0.91,1.00}0.29$\pm$0.45 & \cellcolor[rgb]{0.58,0.58,1.00}6.6$\pm$2.3 & \cellcolor[rgb]{0.67,0.67,1.00}5.69$\pm$2.67 & \cellcolor[rgb]{0.68,0.68,1.00}4.75$\pm$1.12 & \cellcolor[rgb]{0.72,0.72,1.00}12.62$\pm$4.85 \\
Mistral Small & \cellcolor[rgb]{0.98,0.98,1.00}0.07$\pm$0.09 & \cellcolor[rgb]{0.78,0.78,1.00}3.44$\pm$3.15 & \cellcolor[rgb]{0.83,0.83,1.00}2.96$\pm$1.97 & \cellcolor[rgb]{0.77,0.77,1.00}3.51$\pm$1.67 & \cellcolor[rgb]{0.79,0.79,1.00}9.46$\pm$6.76 \\
\midrule
o1 mini & \cellcolor[rgb]{0.52,0.52,1.00}1.49$\pm$1.56 & \cellcolor[rgb]{0.50,0.50,1.00}7.9$\pm$1.81 & \cellcolor[rgb]{0.50,0.50,1.00}8.58$\pm$4.99 & \cellcolor[rgb]{0.50,0.50,1.00}7.53$\pm$1.46 & \cellcolor[rgb]{0.56,0.56,1.00}19.48$\pm$3.56 \\
GPT 4o & \cellcolor[rgb]{0.88,0.88,1.00}0.36$\pm$0.36 & \cellcolor[rgb]{0.51,0.51,1.00}7.75$\pm$1.56 & \cellcolor[rgb]{0.62,0.62,1.00}6.6$\pm$3.97 & \cellcolor[rgb]{0.51,0.51,1.00}7.36$\pm$2.85 & \cellcolor[rgb]{0.66,0.66,1.00}15.29$\pm$4.7 \\
GPT 4o mini & \cellcolor[rgb]{0.95,0.95,1.00}0.16$\pm$0.18 & \cellcolor[rgb]{0.75,0.75,1.00}3.94$\pm$3.97 & \cellcolor[rgb]{0.79,0.79,1.00}3.53$\pm$2.47 & \cellcolor[rgb]{0.74,0.74,1.00}3.88$\pm$1.98 & \cellcolor[rgb]{0.84,0.84,1.00}6.95$\pm$7.85 \\
\midrule
Llama 3.3 70B & \cellcolor[rgb]{0.75,0.75,1.00}0.79$\pm$1.14 & \cellcolor[rgb]{0.80,0.80,1.00}3.22$\pm$2.26 & \cellcolor[rgb]{0.80,0.80,1.00}3.42$\pm$2.86 & \cellcolor[rgb]{0.74,0.74,1.00}3.87$\pm$1.39 & \cellcolor[rgb]{0.86,0.86,1.00}6.16$\pm$4.07 \\
Llama 3.1 70B & \cellcolor[rgb]{0.78,0.78,1.00}0.7$\pm$1.17 & \cellcolor[rgb]{0.67,0.67,1.00}5.23$\pm$2.51 & \cellcolor[rgb]{0.88,0.88,1.00}2.06$\pm$1.33 & \cellcolor[rgb]{0.80,0.80,1.00}3.03$\pm$1.58 & \cellcolor[rgb]{0.87,0.87,1.00}5.76$\pm$3.9 \\
    \bottomrule
    \end{tabular}}
    \label{tab:avg-minatar-evaluation}
\end{table}

For each program, 50 evaluation episodes were performed. All trials resulted in successful programs, i.e., programs that were executable and returned a positive reward. 
\Cref{tab:minatar-evaluation} presents the max. reward of the best program found and \Cref{tab:avg-minatar-evaluation} shows the average reward with the standard deviation of the 10 programs found in each trial. 

In general, larger models outperform their smaller counterparts in average reward, with o1 mini and Claude Sonnet achieving the highest overall performance. This trend is also reflected in the maximum reward: Claude Sonnet generated the best programs for Space Invaders and Asterix, while o1 mini produced the best results for Seaquest and Breakout. GPT 4o found the best performing program for Freeway, although o1 mini slightly outperforms it on average.

In the case of Space Invaders, GPT 4o mini achieved the second best program, though this appears to be a special case, since its average reward is approximately three times lower than that of Claude Sonnet, suggesting that the high performance may be due to chance. The standard deviation across the top 10 programs is often quite high, indicating that multiple trials of program search may be more effective than a single long run.
To explore this hypothesis, we conduct a long running experiment in the Minatar domain, presented in \cref{sec:1000Minatar}, where we also show reward curves to determine when the best programs is found. 
Interestingly, in some cases smaller models outperform their larger counterparts, e.g., Claude Haiku in Seaquest and Freeway and Gemini Flash Lite in Breakout. 
It can be seen from both tables that it is more difficult to find good programs for more complicated games such as Seaquest. 

Looking at the best programs for each LLM in Seaquest, o1 mini is the only model that defines multiple utility functions to check if it can shoot enemies, or implements one-step look-ahead (OSLA) to check if nearby cells are safe to move to. 

In Freeway, all the best performing programs show similar behaviour for each LLM and implement some form of OSLA to check whether it is safe for the chicken to cross the road. GPT 4o's best performing program even uses three look-ahead steps, i.e. it only moves forward if the three cells above the chicken are empty.
\begin{table*}[t]
    \centering
    \caption{The minimum distance of the best program for the Asteroids ship driving experiments for different rotation speeds $\omega$ in degrees per second. $D_{avg}$ shows the average distance of the best programs for the different rotation speeds. For each program, the same set of 5 evaluation tasks were used, each with different target positions and initial states of the ship. The heatmap uses an inverted colour gradient, i.e., values close to zero are darker.}
    \label{tab:vehicle-results}
    \begin{tabular}{lccccccccccc}
    \toprule
    Model & $\omega=10$ & $\omega=20$ & $\omega=30$ & $\omega=40$ & $\omega=50$ & $\omega=60$ & $\omega=70$ & $\omega=80$ & $\omega=90$ & $\omega=100$ & $D_{avg}$ \\
    \midrule
Claude Sonnet & \cellcolor[rgb]{0.84,0.84,1.00}133.36 & \cellcolor[rgb]{0.81,0.81,1.00}97.97 & \cellcolor[rgb]{0.85,0.85,1.00}87.09 & \cellcolor[rgb]{0.71,0.71,1.00}69.18 & \cellcolor[rgb]{0.77,0.77,1.00}70.19 & \cellcolor[rgb]{0.77,0.77,1.00}72.03 & \cellcolor[rgb]{0.86,0.86,1.00}75.75 & \cellcolor[rgb]{0.76,0.76,1.00}66.71 & \cellcolor[rgb]{0.79,0.79,1.00}69.61 & \cellcolor[rgb]{0.75,0.75,1.00}65.67 & \cellcolor[rgb]{0.84,0.84,1.00}80.76 \\
Claude Haiku & \cellcolor[rgb]{0.71,0.71,1.00}101.0 & \cellcolor[rgb]{0.92,0.92,1.00}119.55 & \cellcolor[rgb]{0.88,0.88,1.00}92.98 & \cellcolor[rgb]{0.72,0.72,1.00}71.7 & \cellcolor[rgb]{0.83,0.83,1.00}79.92 & \cellcolor[rgb]{0.74,0.74,1.00}67.01 & \cellcolor[rgb]{0.76,0.76,1.00}62.01 & \cellcolor[rgb]{0.82,0.82,1.00}74.83 & \cellcolor[rgb]{0.78,0.78,1.00}69.19 & \cellcolor[rgb]{0.82,0.82,1.00}75.37 & \cellcolor[rgb]{0.84,0.84,1.00}81.36 \\
\midrule
Gemini Flash & \cellcolor[rgb]{1.00,1.00,1.00}171.79 & \cellcolor[rgb]{0.87,0.87,1.00}109.88 & \cellcolor[rgb]{0.81,0.81,1.00}81.12 & \cellcolor[rgb]{0.80,0.80,1.00}85.61 & \cellcolor[rgb]{0.76,0.76,1.00}69.4 & \cellcolor[rgb]{0.79,0.79,1.00}74.91 & \cellcolor[rgb]{0.86,0.86,1.00}76.53 & \cellcolor[rgb]{0.81,0.81,1.00}74.07 & \cellcolor[rgb]{0.77,0.77,1.00}67.49 & \cellcolor[rgb]{0.81,0.81,1.00}74.02 & \cellcolor[rgb]{0.89,0.89,1.00}88.48 \\
Gemini Lite & \cellcolor[rgb]{0.99,0.99,1.00}169.27 & \cellcolor[rgb]{0.85,0.85,1.00}106.47 & \cellcolor[rgb]{0.87,0.87,1.00}91.18 & \cellcolor[rgb]{0.72,0.72,1.00}72.45 & \cellcolor[rgb]{0.71,0.71,1.00}62.3 & \cellcolor[rgb]{0.78,0.78,1.00}73.63 & \cellcolor[rgb]{0.72,0.72,1.00}57.69 & \cellcolor[rgb]{1.00,1.00,1.00}101.62 & \cellcolor[rgb]{1.00,1.00,1.00}100.37 & \cellcolor[rgb]{0.77,0.77,1.00}68.42 & \cellcolor[rgb]{0.90,0.90,1.00}90.34 \\
\midrule
Mistral Large & \cellcolor[rgb]{0.76,0.76,1.00}113.69 & \cellcolor[rgb]{0.70,0.70,1.00}76.54 & \cellcolor[rgb]{0.82,0.82,1.00}82.5 & \cellcolor[rgb]{0.70,0.70,1.00}68.81 & \cellcolor[rgb]{0.76,0.76,1.00}68.42 & \cellcolor[rgb]{0.85,0.85,1.00}83.33 & \cellcolor[rgb]{0.71,0.71,1.00}55.62 & \cellcolor[rgb]{0.74,0.74,1.00}63.58 & \cellcolor[rgb]{0.73,0.73,1.00}61.58 & \cellcolor[rgb]{0.69,0.69,1.00}57.04 & \cellcolor[rgb]{0.79,0.79,1.00}73.11 \\
Mistral Small & \cellcolor[rgb]{0.89,0.89,1.00}144.33 & \cellcolor[rgb]{0.77,0.77,1.00}90.16 & \cellcolor[rgb]{0.75,0.75,1.00}71.24 & \cellcolor[rgb]{0.89,0.89,1.00}100.13 & \cellcolor[rgb]{0.69,0.69,1.00}57.9 & \cellcolor[rgb]{0.70,0.70,1.00}60.84 & \cellcolor[rgb]{0.80,0.80,1.00}68.51 & \cellcolor[rgb]{0.70,0.70,1.00}58.6 & \cellcolor[rgb]{0.70,0.70,1.00}56.92 & \cellcolor[rgb]{0.69,0.69,1.00}56.53 & \cellcolor[rgb]{0.81,0.81,1.00}76.52 \\
\midrule
o1 mini & \cellcolor[rgb]{0.80,0.80,1.00}123.64 & \cellcolor[rgb]{0.83,0.83,1.00}101.49 & \cellcolor[rgb]{0.84,0.84,1.00}86.57 & \cellcolor[rgb]{0.76,0.76,1.00}77.88 & \cellcolor[rgb]{0.86,0.86,1.00}83.49 & \cellcolor[rgb]{0.73,0.73,1.00}64.92 & \cellcolor[rgb]{0.88,0.88,1.00}79.07 & \cellcolor[rgb]{0.79,0.79,1.00}71.62 & \cellcolor[rgb]{0.70,0.70,1.00}57.42 & \cellcolor[rgb]{0.79,0.79,1.00}71.16 & \cellcolor[rgb]{0.84,0.84,1.00}81.73 \\
GPT 4o & \cellcolor[rgb]{0.82,0.82,1.00}127.58 & \cellcolor[rgb]{0.82,0.82,1.00}99.21 & \cellcolor[rgb]{0.80,0.80,1.00}78.93 & \cellcolor[rgb]{0.95,0.95,1.00}111.06 & \cellcolor[rgb]{0.73,0.73,1.00}64.91 & \cellcolor[rgb]{0.96,0.96,1.00}99.81 & \cellcolor[rgb]{1.00,1.00,1.00}95.28 & \cellcolor[rgb]{0.87,0.87,1.00}83.24 & \cellcolor[rgb]{0.82,0.82,1.00}74.2 & \cellcolor[rgb]{0.72,0.72,1.00}61.57 & \cellcolor[rgb]{0.90,0.90,1.00}89.58 \\
GPT 4o mini & \cellcolor[rgb]{0.89,0.89,1.00}144.86 & \cellcolor[rgb]{1.00,1.00,1.00}134.72 & \cellcolor[rgb]{0.96,0.96,1.00}104.54 & \cellcolor[rgb]{0.95,0.95,1.00}111.67 & \cellcolor[rgb]{0.90,0.90,1.00}89.83 & \cellcolor[rgb]{0.92,0.92,1.00}94.26 & \cellcolor[rgb]{0.92,0.92,1.00}84.0 & \cellcolor[rgb]{0.96,0.96,1.00}96.41 & \cellcolor[rgb]{0.96,0.96,1.00}95.25 & \cellcolor[rgb]{0.97,0.97,1.00}96.99 & \cellcolor[rgb]{1.00,1.00,1.00}105.25 \\
\midrule
Llama 3.3 70B & \cellcolor[rgb]{0.92,0.92,1.00}152.17 & \cellcolor[rgb]{0.86,0.86,1.00}106.87 & \cellcolor[rgb]{1.00,1.00,1.00}111.51 & \cellcolor[rgb]{1.00,1.00,1.00}119.46 & \cellcolor[rgb]{1.00,1.00,1.00}105.09 & \cellcolor[rgb]{1.00,1.00,1.00}106.36 & \cellcolor[rgb]{0.87,0.87,1.00}77.21 & \cellcolor[rgb]{0.97,0.97,1.00}97.09 & \cellcolor[rgb]{0.83,0.83,1.00}75.86 & \cellcolor[rgb]{1.00,1.00,1.00}101.72 & \cellcolor[rgb]{1.00,1.00,1.00}105.33 \\
Llama 3.1 70B & \cellcolor[rgb]{0.85,0.85,1.00}135.31 & \cellcolor[rgb]{0.80,0.80,1.00}96.67 & \cellcolor[rgb]{0.88,0.88,1.00}92.75 & \cellcolor[rgb]{0.72,0.72,1.00}72.29 & \cellcolor[rgb]{0.80,0.80,1.00}74.85 & \cellcolor[rgb]{0.71,0.71,1.00}62.51 & \cellcolor[rgb]{0.76,0.76,1.00}62.15 & \cellcolor[rgb]{0.75,0.75,1.00}64.76 & \cellcolor[rgb]{0.71,0.71,1.00}58.73 & \cellcolor[rgb]{0.73,0.73,1.00}62.49 & \cellcolor[rgb]{0.82,0.82,1.00}78.25 \\
    \bottomrule
    \end{tabular}
\end{table*}

For Breakout, the best programs are all able to locate the ball and also determine the direction in which the ball is moving. Claude Sonnet and o1 mini even check whether the ball bounces off walls. The best performing program of o1 mini simulates the movement of the ball until it is in the same row as the paddle.

In Asterix, the best performing programs with a reward over 10 prioritise reaching the gold while checking whether it is moving towards an enemy and, if so, avoiding the enemy. The mediocre performing models often prioritize the gold without checking if there are enemies nearby. Claude Sonnet even takes into account the trail of the enemy to check if the chosen action is safe.  

In Space Invader, the good programs with a reward of over 20 correctly locate the enemy bullets, the aliens and the cannon and then use threat detection to check whether the enemy bullets need to be dodged before the enemies themselves are shot. The programs with a score below 20 do not anticipate the movement of aliens or prioritise shooting enemies over avoiding enemy bullets.

Overall, it can be said that in the Minatar games larger models on average show more sophisticated behaviour in the programs, but as can be seen with Claude Haiku or Gemini Flash Lite, this is not always the case, since for some games they achieve a better reward. Currently, only a very simple prompting strategy is used, which already gives comparable results to some of the baselines reported in \cite{minatar} or even outperforms all baselines in the case of Breakout. Using more complicated prompting strategies, such as Chain of Thought \cite{wei2022chain} or adding a crossover operator could lead to improvements in the programs found. 

\subsection{Vehicle Driving}
The task is to pilot an Asteroids-style spaceship from its start state to the target, where it should rest until the end of the episode.  Each episode is 101 steps.   At each step, there are 4 discrete actions: NO\_OP, THRUST, ROTATE\_LEFT, ROTATE\_RIGHT.  We experimented with vehicle physics in order to make an interesting challenge.  Drag is set to be low, which leads to a high risk of overshooting the target unless countermeasures are taken.  At each step, the agent is given an observation of the ship state and the position of the target.  

The prompt includes some helper classes and functions, including a Vector2d class and
the Asteroids ship, as well as a Vehicle superclass.
In addition, we add strong hints to make the problem solvable for LLMs, which are summarized as follows\footnote{The complete prompt is given in the code repository.}:
\begin{itemize}
    \item Best solved using search algorithms: try One Step Lookahead, Monte Carlo Tree Search or Rolling Horizon Evolution.
    \item Try using a heuristic function that values facing towards the target as well as being close to the target.
    \item Try using Macro-actions - e.g. simply repeating each action a number of times.
\end{itemize}

\Cref{tab:vehicle-results} shows the results of the driving experiments for different rotation speeds $\omega$ to adjust the difficulty level of steering the asteroid ship. The numbers are the minimum distance achieved by the best program for five evaluation episodes. $D_{avg}$ is the average of all distances for each LLM.
We omit the number of successful iterations because no LLM managed to stop the asteroid ship at the target position in all five evaluation episodes. A program was considered successful if it could consistently stop the vehicle within a specified tolerance $t$ across all evaluation episodes. In our experiments, the synthesized programs succeeded in stopping the vehicle in only one or two episodes within a tolerance of $t=10$, and thus no program qualified as successful. The best models overall are the two Mistral LLMs, which together achieve the best programs for 9 out of 10 rotation speeds. Only for $\omega=10$ did another LLM (Claude Haiku) find the best program. In terms of $D_{avg}$, larger models generally outperformed their smaller counterparts but usually only by a small margin. As no LLM successfully solved the problem with a simple prompting strategy in this experiment, we consider it a compelling challenge for future research.

\begin{figure*}[ht]
    \centering
    \begin{subfigure}[t]{0.13\textwidth}
        \includegraphics[width=\textwidth]{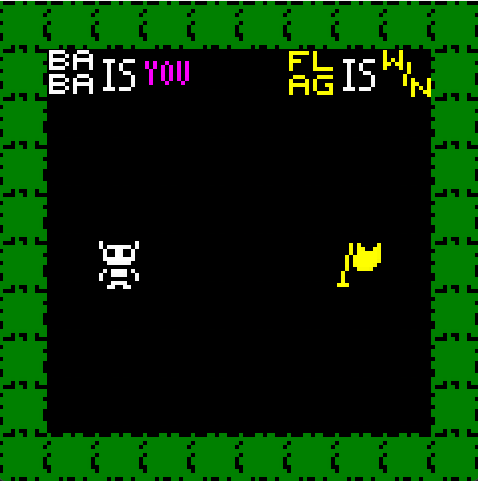}
        \caption{Level 1}
    \end{subfigure}
    \begin{subfigure}[t]{0.13\textwidth}
        \includegraphics[width=\textwidth]{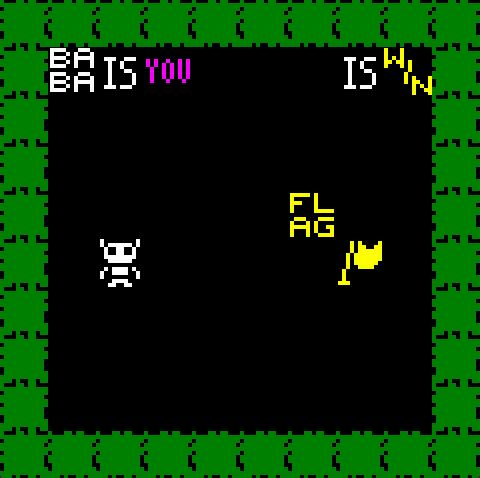}
        \caption{Level 2}
    \end{subfigure}
    \begin{subfigure}[t]{0.13\textwidth}
        \includegraphics[width=\textwidth]{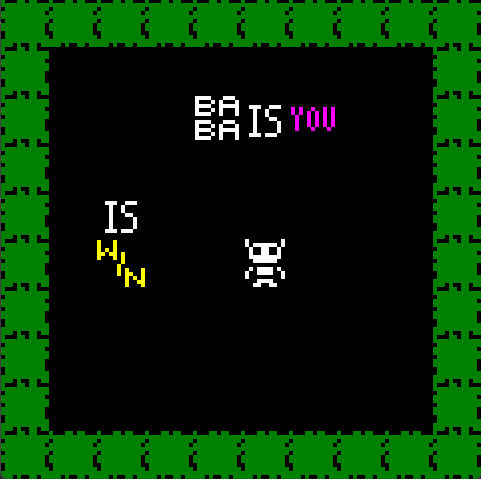}
        \caption{Level 3}
    \end{subfigure}
    \begin{subfigure}[t]{0.13\textwidth}
        \includegraphics[width=\textwidth]{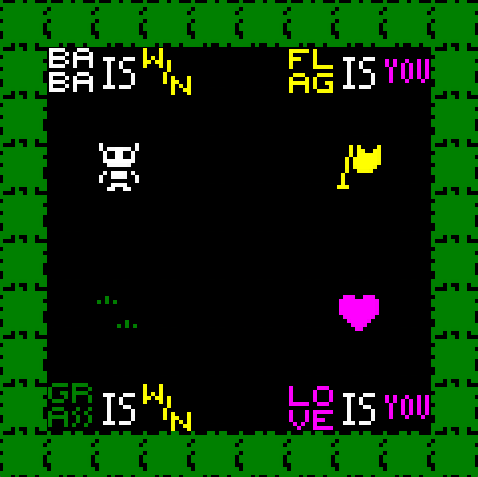}
        \caption{Level 4}
    \end{subfigure}
    \begin{subfigure}[t]{0.13\textwidth}
        \includegraphics[width=\textwidth]{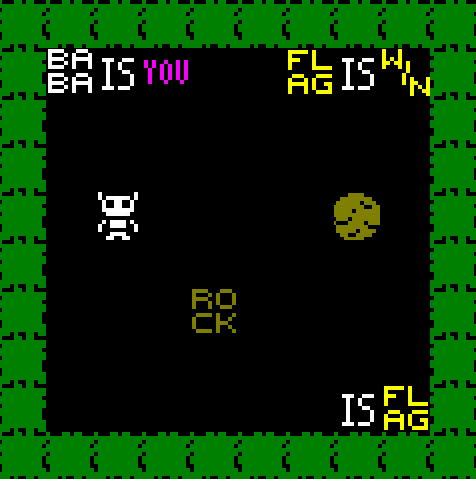}
        \caption{Level 5}
    \end{subfigure}
    \vspace{1em}
    
    \begin{subfigure}[t]{0.13\textwidth}
        \includegraphics[width=\textwidth]{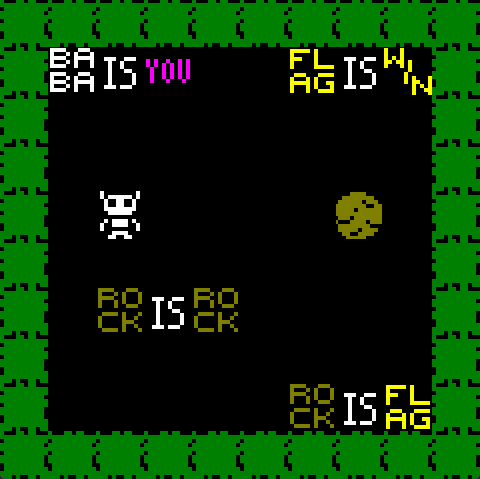}
        \caption{Level 6}
    \end{subfigure}
    \begin{subfigure}[t]{0.13\textwidth}
        \includegraphics[width=\textwidth]{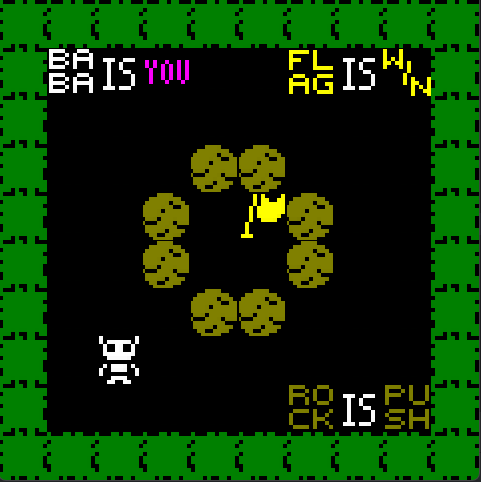}
        \caption{Level 7}
    \end{subfigure}
    \begin{subfigure}[t]{0.13\textwidth}
        \includegraphics[width=\textwidth]{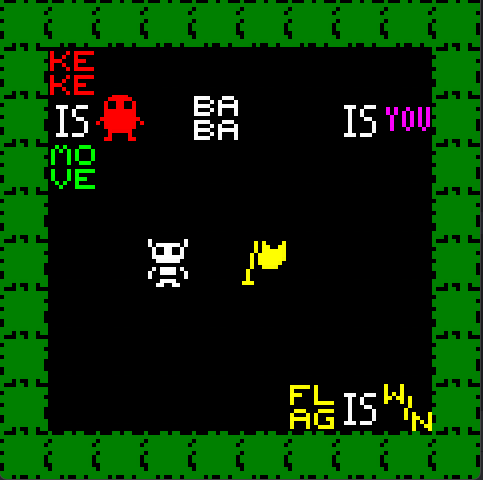}
        \caption{Level 8}
    \end{subfigure}
    \begin{subfigure}[t]{0.13\textwidth}
        \includegraphics[width=\textwidth]{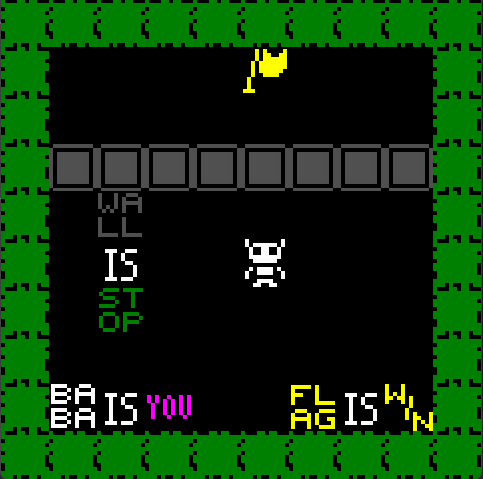}
        \caption{Level 9}
    \end{subfigure}
    \begin{subfigure}[t]{0.13\textwidth}
        \includegraphics[width=\textwidth]{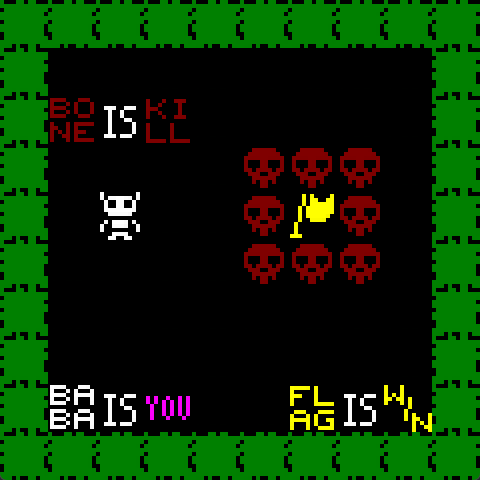}
        \caption{Level 10}
    \end{subfigure}
    
    \caption{Demo levels used for the evaluation of LLM capabilities in the Baba is You domain.}
    \label{fig:baba-is-you-levels}
\end{figure*}

\begin{table*}[ht]
    \centering
    \caption{Highest reward per language model and level (with number of successful trials per level). The color of the heatmap is based on the successful trials, as it is more important to know how many times the LLMs were able to solve a level out of all 10 trials or if the trials were solved randomly.}
    \label{tab:results-baba-is-you}
    \begin{tabular}{@{}lccccccccccc@{}}
        \toprule
        Model & \textbf{Level 1} & \textbf{Level 2} & \textbf{Level 3} & \textbf{Level 4} & \textbf{Level 5} & \textbf{Level 6} & \textbf{Level 7} & \textbf{Level 8} & \textbf{Level 9} & \textbf{Level 10} &\textbf{\#Levels} \\
         & & & & & & & & & & &\textbf{solved}\\
        \midrule
        Claude Sonnet& \cellcolor[rgb]{0.50,0.50,1.00} 95 (10) & \cellcolor[rgb]{0.75,0.75,1.00} 89 (5) & \cellcolor[rgb]{1.00,1.00,1.00} 0.0 (0) & \cellcolor[rgb]{0.50,0.50,1.00} 95 (10) & \cellcolor[rgb]{1.00,1.00,1.00} 0.0 (0) & \cellcolor[rgb]{0.95,0.95,1.00} 91 (1) & \cellcolor[rgb]{0.50,0.50,1.00} 94 (10) & \cellcolor[rgb]{0.50,0.50,1.00} 95 (10) & \cellcolor[rgb]{0.95,0.95,1.00} 90 (1) & \cellcolor[rgb]{0.85,0.85,1.00} 92 (3) & 8 \\
        Claude Haiku & \cellcolor[rgb]{0.50,0.50,1.00} 95 (10) & \cellcolor[rgb]{1.00,1.00,1.00} 0.0 (0) & \cellcolor[rgb]{1.00,1.00,1.00} 0.0 (0) & \cellcolor[rgb]{0.50,0.50,1.00} 95 (10) & \cellcolor[rgb]{1.00,1.00,1.00} 0.0 (0) & \cellcolor[rgb]{1.00,1.00,1.00} 0.0 (0) & \cellcolor[rgb]{0.50,0.50,1.00} 94 (10) & \cellcolor[rgb]{0.50,0.50,1.00} 95 (10) & \cellcolor[rgb]{1.00,1.00,1.00} 0.0 (0) & \cellcolor[rgb]{1.00,1.00,1.00} 0.0 (0) & 4 \\
        \midrule
        Gemini Flash     & \cellcolor[rgb]{0.50,0.50,1.00} 95 (10) & \cellcolor[rgb]{1.00,1.00,1.00} 0.0 (0) & \cellcolor[rgb]{1.00,1.00,1.00} 0.0 (0) & \cellcolor[rgb]{0.90,0.90,1.00} 95 (2) & \cellcolor[rgb]{1.00,1.00,1.00} 0.0 (0) & \cellcolor[rgb]{1.00,1.00,1.00} 0.0 (0) & \cellcolor[rgb]{0.65,0.65,1.00} 94 (7) & \cellcolor[rgb]{0.55,0.55,1.00} 95 (9) & \cellcolor[rgb]{1.00,1.00,1.00} 0.0 (0) & \cellcolor[rgb]{1.00,1.00,1.00} 0.0 (0) & 4 \\
        Gemini Lite& \cellcolor[rgb]{0.50,0.50,1.00} 95 (10) & \cellcolor[rgb]{1.00,1.00,1.00} 0.0 (0) & \cellcolor[rgb]{1.00,1.00,1.00} 0.0 (0) & \cellcolor[rgb]{0.90,0.90,1.00} 95 (2) & \cellcolor[rgb]{1.00,1.00,1.00} 0.0 (0) & \cellcolor[rgb]{1.00,1.00,1.00} 0.0 (0) & \cellcolor[rgb]{0.65,0.65,1.00} 94 (7) & \cellcolor[rgb]{0.55,0.55,1.00} 95 (9) & \cellcolor[rgb]{1.00,1.00,1.00} 0.0 (0) & \cellcolor[rgb]{1.00,1.00,1.00} 0.0 (0) & 4 \\
        \midrule
        Mistral Large    & \cellcolor[rgb]{0.70,0.70,1.00} 95 (6) & \cellcolor[rgb]{0.95,0.95,1.00} 89 (1) & \cellcolor[rgb]{1.00,1.00,1.00} 0.0 (0) & \cellcolor[rgb]{0.95,0.95,1.00} 72 (1) & \cellcolor[rgb]{1.00,1.00,1.00} 0.0 (0) & \cellcolor[rgb]{1.00,1.00,1.00} 0.0 (0) & \cellcolor[rgb]{0.55,0.55,1.00} 94 (9) & \cellcolor[rgb]{0.65,0.65,1.00} 95 (7) & \cellcolor[rgb]{1.00,1.00,1.00} 0.0 (0) & \cellcolor[rgb]{1.00,1.00,1.00} 0.0 (0) & 5 \\
        Mistral Small    &\cellcolor[rgb]{0.70,0.70,1.00} 95 (6) & \cellcolor[rgb]{1.00,1.00,1.00} 0.0 (0) & \cellcolor[rgb]{1.00,1.00,1.00} 0.0 (0) & \cellcolor[rgb]{0.95,0.95,1.00} 63 (1) & \cellcolor[rgb]{1.00,1.00,1.00} 0.0 (0) & \cellcolor[rgb]{1.00,1.00,1.00} 0.0 (0) & \cellcolor[rgb]{0.90,0.90,1.00} 94 (2) & \cellcolor[rgb]{0.60,0.60,1.00} 95 (8) & \cellcolor[rgb]{0.90,0.90,1.00} 64 (2) & \cellcolor[rgb]{1.00,1.00,1.00} 0.0 (0)& 5 \\
        \midrule
        o1 mini          & \cellcolor[rgb]{0.50,0.50,1.00} 95 (10) & \cellcolor[rgb]{0.90,0.90,1.00} 89 (2) & \cellcolor[rgb]{1.00,1.00,1.00} 0.0 (0) & \cellcolor[rgb]{0.50,0.50,1.00} 95 (10) & \cellcolor[rgb]{1.00,1.00,1.00} 0.0 (0) & \cellcolor[rgb]{1.00,1.00,1.00} 0.0 (0) & \cellcolor[rgb]{0.50,0.50,1.00} 94 (10) & \cellcolor[rgb]{0.50,0.50,1.00} 95 (10) & \cellcolor[rgb]{0.95,0.95,1.00} 82 (1) & \cellcolor[rgb]{0.75,0.75,1.00} 92 (5)  & 7 \\
        GPT 4o           & \cellcolor[rgb]{0.50,0.50,1.00} 95 (10) & \cellcolor[rgb]{0.95,0.95,1.00} 89 (1) & \cellcolor[rgb]{1.00,1.00,1.00} 0.0 (0) & \cellcolor[rgb]{0.55,0.55,1.00} 95 (9) & \cellcolor[rgb]{1.00,1.00,1.00} 0.0 (0) & \cellcolor[rgb]{1.00,1.00,1.00} 0.0 (0) & \cellcolor[rgb]{0.60,0.60,1.00} 94 (8) & \cellcolor[rgb]{0.55,0.55,1.00} 95 (9) & \cellcolor[rgb]{1.00,1.00,1.00} 0.0 (0) & \cellcolor[rgb]{0.95,0.95,1.00} 92 (1)& 5 \\
        GPT 4o mini      & \cellcolor[rgb]{0.60,0.60,1.00} 95 (8) & \cellcolor[rgb]{1.00,1.00,1.00} 0.0 (0) & \cellcolor[rgb]{1.00,1.00,1.00} 0.0 (0) & \cellcolor[rgb]{1.00,1.00,1.00} 0.0 (0) & \cellcolor[rgb]{1.00,1.00,1.00} 0.0 (0) & \cellcolor[rgb]{1.00,1.00,1.00} 0.0 (0) & \cellcolor[rgb]{0.85,0.85,1.00} 90 (3) & \cellcolor[rgb]{0.80,0.80,1.00} 95 (4) & \cellcolor[rgb]{1.00,1.00,1.00} 0.0 (0) & \cellcolor[rgb]{1.00,1.00,1.00} 0.0 (0) & 3 \\
        \midrule
        Llama 3.3 70B    &\cellcolor[rgb]{0.50,0.50,1.00} 95 (10) & \cellcolor[rgb]{1.00,1.00,1.00} 0.0 (0) & \cellcolor[rgb]{1.00,1.00,1.00} 0.0 (0) & \cellcolor[rgb]{0.85,0.85,1.00} 95 (3) & \cellcolor[rgb]{1.00,1.00,1.00} 0.0 (0) & \cellcolor[rgb]{1.00,1.00,1.00} 0.0 (0) & \cellcolor[rgb]{0.70,0.70,1.00} 94 (6) & \cellcolor[rgb]{0.70,0.70,1.00} 95 (6) & \cellcolor[rgb]{1.00,1.00,1.00} 0.0 (0) & \cellcolor[rgb]{1.00,1.00,1.00} 0.0 (0) &  4 \\
        Llama 3.1 70B    & \cellcolor[rgb]{0.50,0.50,1.00} 95 (10) & \cellcolor[rgb]{1.00,1.00,1.00} 0.0 (0) & \cellcolor[rgb]{1.00,1.00,1.00} 0.0 (0) & \cellcolor[rgb]{0.90,0.90,1.00} 95 (2) & \cellcolor[rgb]{1.00,1.00,1.00} 0.0 (0) & \cellcolor[rgb]{0.95,0.95,1.00} 51 (1) & \cellcolor[rgb]{0.80,0.80,1.00} 94 (4) & \cellcolor[rgb]{0.70,0.70,1.00} 95 (6) & \cellcolor[rgb]{1.00,1.00,1.00} 0.0 (0) & \cellcolor[rgb]{1.00,1.00,1.00} 0.0 (0) & 5 \\
        \bottomrule
    \end{tabular}
\end{table*}

\subsection{Baba is You}

\textit{Baba is you} is a complex puzzle game in which the player manipulates a 2D grid environment to reach a given goal. The environment consists of word blocks and corresponding entities that can be pushed around. By placing word blocks next to each other, rules can be formed. These rules are active as long as the given word block sequence remains intact. This way, players can change how objects behave, which objects they control, or which conditions must be satisfied to win.

For our experiments, we used a Python version\footnote{\url{https://github.com/ADockhorn/Keke-AI-PY}} of the Keke is You AI framework~\cite{charity2022keke}. For this domain, we prompted the LLMs to provide a policy, giving a short description of the game and the initial state of the level (the complete prompt is in the repository). Similar to the Minatar experiments, the state is converted into a text description. The function to be written should use the current state as input and return a movement direction or the command for waiting a turn.  Each episode ends after 100 actions or once the win condition is fulfilled. A reward is awarded based on the maximum number of actions (100) minus the number of steps taken. Thus, the return can be maximized by finishing the level as fast as possible. Each level can be solved in less than 20 actions.

In our tests, we queried the agent to solve 10 simple demo levels (see \Cref{fig:baba-is-you-levels}). Each of the levels focuses on one or more key mechanics of the framework such as rule interpretation (levels 1-10), rule creation (levels 2, 3, 5) or destruction (levels 6, 8, 9, 10), and object manipulation (level 7).
\Cref{tab:results-baba-is-you} shows the results of our comparison of the LLM models' capabilities. The number of successful iterations is shown in brackets.

All agents were able to solve at least 3 out of 10 levels, with Claude 3.5 Sonnet being the only model able to solve 8. For Claude and GPT, models of the same vendor with a higher number of parameters were able to solve more levels. For the Llama models, the 3.1 version solved one more level than the 3.3 model. In case of Gemini and Mistral, both the small and large models performed about the same. Tested models were mostly successful in interpreting existing rules.
As can be seen, some levels are rarely solved by any model. Creating or destroying rules and thus modifying the logic of our game world has proven difficult for all models. Many models failed in solving levels 2 and 3, which require rule creation, and levels 9 and 10, which require a rule's destruction to finish the puzzle. Slight differences in the observed success rate could be due to the low number of repetitions per level, resulting in sampling errors: levels 2, 6, 9, and 10, which are rarely solved at all, could be affected by this. Chain of thought prompting \cite{wei2022chain} may help in overcoming these more complex planning tasks.

\subsection{Procedural Content Generation}

\begin{figure}
    \centering
    \includegraphics[width=0.95\linewidth]{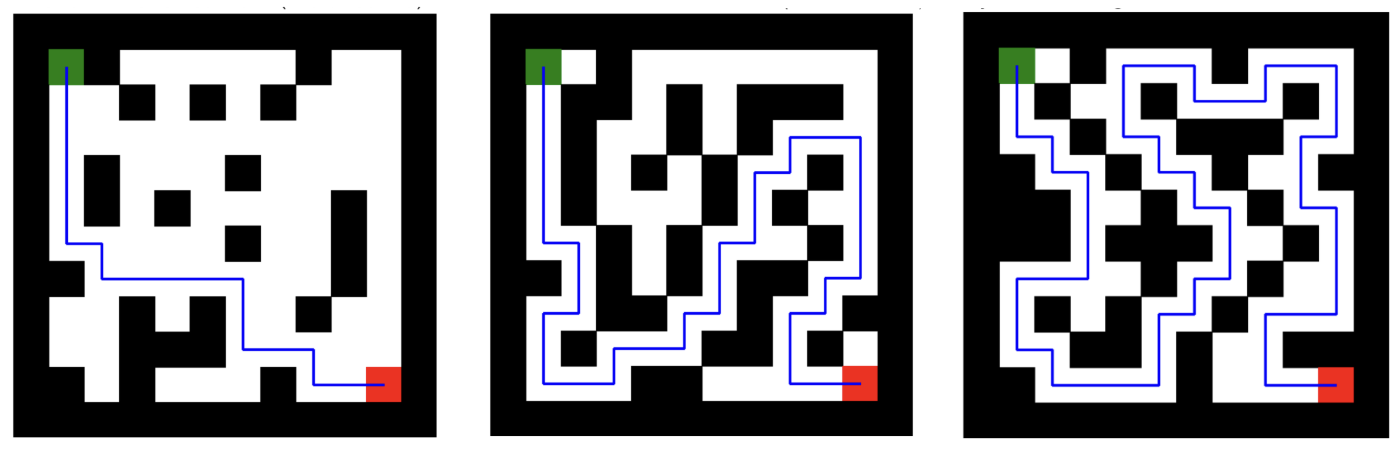}
    \caption{Three generated mazes aiming to optimise the longest
    shortest path objective.  Left: (score 18) example from a simple LLM generated algorithm setting wall cells
    with a fixed probability.  Middle: (score 38) example a more sophisticated LLM algorithm involving recursion and a shortest path algorithm.  Right: (score 54) example from an evolutionary algorithm directly optimising for the objective function.}
    \label{fig:random-maze}
\end{figure}
PCG is a widely studied area in game research \cite{pcg-book, pcgml}. In this experiment, we explored whether LLMs can synthesize Python functions capable of generating diverse game content. To assess this in a simple scenario, we tasked the LLMs with creating functions that produce random mazes adhering to specific design objectives.

The prompt advised the LLMs to use the longest shortest path objective
to guide the maze generation process. This objective encourages intricate
and interesting mazes.  Most of the generated code ignored the hint
and instead coded overly simple algorithms, placing corridors and walls in 
each cell with a given probability while usually ensuring that the start and end points were not on wall cells.  An example generated maze is shown in the left of figure~\ref{fig:random-maze}.  
Occasionally, a better algorithm was produced that mixed randomness, recursion and graph search in ways we have not fully analysed. These algorithms sometimes produced mazes with no path between start and end, resulting in a score of -1.  When they worked, they often produced reasonable mazes such as the one shown in the middle of Figure~\ref{fig:random-maze}.  The LLMs failed to find an algorithm as effective as an evolutionary algorithm applied to directly solve the objective.  A sample maze from such an algorithm is shown on the right of the figure.

Note that here we are evaluating the effectiveness of the algorithms in meeting
the specified objective, which is to produce mazes with the longest shortest path between start and end. Depending on the application, this could be a poor objective to maximise, with the best mazes having a mid-ranking score, such as the central maze in Figure~\ref{fig:random-maze}.

\begin{table}[t]
    \centering
    \caption{Maze generation LLM results. $D_{max}$ is the maximum distance of the shortest path of the generated mazes returned by the best program of all trials and $D_{avg}\pm\sigma$ is the average distance with the standard deviation of the 10 programs found after all trials. Each program generates five mazes for evaluating $D_{max}$ and $D_{avg}\pm\sigma$.}  
    \label{tab:mazegen-evaluation}
    \begin{tabular}{lccccc}
    \toprule
    Model & $D_{max}$ & $D_{avg}\pm\sigma$ \\
    \midrule
Claude Sonnet & \cellcolor[rgb]{0.80,0.80,1.00}18.0 & \cellcolor[rgb]{0.80,0.80,1.00}8.98$\pm$3.79 \\
Claude Haiku & \cellcolor[rgb]{0.67,0.67,1.00}29.4 & \cellcolor[rgb]{0.65,0.65,1.00}15.4$\pm$8.73 \\
\midrule
Gemini Flash & \cellcolor[rgb]{0.81,0.81,1.00}17.0 & \cellcolor[rgb]{0.64,0.64,1.00}15.7$\pm$1.28 \\
Gemini Lite & \cellcolor[rgb]{0.74,0.74,1.00}23.2 & \cellcolor[rgb]{0.67,0.67,1.00}14.36$\pm$5.46 \\
\midrule
Mistral Large & \cellcolor[rgb]{0.88,0.88,1.00}10.8 & \cellcolor[rgb]{0.90,0.90,1.00}4.2$\pm$3.3 \\
Mistral Small & \cellcolor[rgb]{0.88,0.88,1.00}10.8 & \cellcolor[rgb]{0.90,0.90,1.00}4.6$\pm$3.65 \\
\midrule
o1 mini & \cellcolor[rgb]{0.50,0.50,1.00}44.4 & \cellcolor[rgb]{0.58,0.58,1.00}18.56$\pm$15.92 \\
GPT 4o & \cellcolor[rgb]{0.62,0.62,1.00}33.4 & \cellcolor[rgb]{0.58,0.58,1.00}18.48$\pm$11.66 \\
GPT 4o mini & \cellcolor[rgb]{0.62,0.62,1.00}33.4 & \cellcolor[rgb]{0.50,0.50,1.00}22.02$\pm$11.76 \\
\midrule
Llama 3.3 70B & \cellcolor[rgb]{0.83,0.83,1.00}14.8 & \cellcolor[rgb]{0.81,0.81,1.00}8.3$\pm$3.75 \\
Llama 3.1 70B & \cellcolor[rgb]{0.87,0.87,1.00}11.4 & \cellcolor[rgb]{0.85,0.85,1.00}6.76$\pm$2.67 \\
    \bottomrule
    \end{tabular}
\end{table}

\Cref{tab:mazegen-evaluation} presents the results of the maze generation experiment. Unlike the previous tasks, smaller models perform in line with, or even outperform, their larger counterparts. Among the larger models, only GPT-4o achieves performance on par with the smaller models, although it performs slightly worse in terms of $D_{avg}$. Mistral Large also performs similarly to Mistral Small, but both models cannot keep up with models from other LLM providers. As with \Cref{tab:avg-minatar-evaluation}, the high standard deviation of $D_{avg}$ in this experiment  supports the additional long-running experiments in \Cref{sec:1000Minatar}.

The reasoning model o1-mini outperforms both variants of GPT-4o with respect to $D_{max}$, but falls short compared to GPT-4o mini in terms of average distance. Overall, most models struggle with this task, with the exception of the OpenAI models and Claude Haiku.
Understanding why larger language models underperform in this domain remains an open question, but is beyond the scope of this paper. 

\subsection{Python Code Evaluation}
\Cref{tab:python-eval} shows the summary statistics of the synthesized Python code for the Minatar, Baba is You, maze generation and vehicle driving experiments, including the costs incurred. On average, the larger models do better than their small counterpart. In the case of Llama the newer 3.3 model beats the 3.1 model in terms of the successful iterations and trials. The successful iterations are, however, quite similar for all LLMs. The Mistral models have by far the worst percentage of executable programs but are still on the 4th and 6th rank in terms of overall performance, which indicates that the percentage of executable programs are not a good proxy for selecting an LLM to generate game-related Python code. o1 mini is on average the best model on all Python domains, and is also the only reasoning model in this benchmark. This indicates that reasoning models have an advantage in generating game-specific code. A more fine-grained ranking of LLMs for each domain is presented in \Cref{tab:global_ranking} in \Cref{sec:practical}, where we discuss the experiments in general and also give some practical recommendations.

\begin{table}[]
    \centering
    \caption{The overall evaluation of the synthesized Python code. Cost is the total cost of all 2600 iterations for the LLM (100 per game) (the Llama 3.1 70B model is provided for free by Google Cloud as it is currently in public preview). S.Iter. is the percentage of iterations that resulted in working code and S.Trl the percentage of trials (each consisting of 10 iterations). Only programs that returned a positive reward were considered for S.Iter and S.Trl. Exec. Programs is the percentage of all generated programs that the Python interpreter could run. For each of the four domains each LLM is ranked from 1 to 11 based on the performance of the best result for the domain. This rank is averaged and used to assign an overall Rank.}
    \label{tab:python-eval}
    \begin{tabular}{lrcccc}
    \toprule
    Model & Cost (\$) & S.Iter & S.Trl & Exec. Programs & Rank \\
    \midrule
Claude Sonnet & 80.75 & \cellcolor[rgb]{0.68,0.68,1.00}63.27 & \cellcolor[rgb]{0.60,0.60,1.00}80.77 & \cellcolor[rgb]{0.58,0.58,1.00}83.23 &  \cellcolor[rgb]{0.55,0.55,1.00}2nd \\
Claude Haiku & 17.34 & \cellcolor[rgb]{0.69,0.69,1.00}61.35 & \cellcolor[rgb]{0.62,0.62,1.00}76.92 & \cellcolor[rgb]{0.57,0.57,1.00}86.35  & \cellcolor[rgb]{0.70,0.70,1.00}5 \\
\midrule
Gemini Flash & 2.56 & \cellcolor[rgb]{0.69,0.69,1.00}62.62 & \cellcolor[rgb]{0.64,0.64,1.00}72.31 & \cellcolor[rgb]{0.52,0.52,1.00}95.85   &  \cellcolor[rgb]{0.80,0.80,1.00}7 \\
Gemini Lite & 1.30 & \cellcolor[rgb]{0.69,0.69,1.00}61.15 & \cellcolor[rgb]{0.64,0.64,1.00}71.54 & \cellcolor[rgb]{0.53,0.53,1.00}94.28     & \cellcolor[rgb]{0.95,0.95,1.00}10 \\
\midrule
Mistral Large & 43.57 & \cellcolor[rgb]{0.70,0.70,1.00}60.08 & \cellcolor[rgb]{0.65,0.65,1.00}70.77 & \cellcolor[rgb]{0.62,0.62,1.00}75.37 & \cellcolor[rgb]{0.65,0.65,1.00}4 \\
Mistral Small & 2.06 & \cellcolor[rgb]{0.70,0.70,1.00}59.65 & \cellcolor[rgb]{0.66,0.66,1.00}68.85 & \cellcolor[rgb]{0.63,0.63,1.00}74.7   & \cellcolor[rgb]{0.75,0.75,1.00}6 \\
\midrule
o1 mini & 35.87 & \cellcolor[rgb]{0.69,0.69,1.00}62.88 & \cellcolor[rgb]{0.60,0.60,1.00}80.0 & \cellcolor[rgb]{0.55,0.55,1.00}90.93        & \cellcolor[rgb]{0.50,0.50,1.00}1st \\
GPT 4o & 36.11 & \cellcolor[rgb]{0.69,0.69,1.00}62.19 & \cellcolor[rgb]{0.62,0.62,1.00}76.15 & \cellcolor[rgb]{0.54,0.54,1.00}92.09        & \cellcolor[rgb]{0.60,0.60,1.00}3rd \\
GPT 4o mini & 2.18 & \cellcolor[rgb]{0.69,0.69,1.00}62.0 & \cellcolor[rgb]{0.66,0.66,1.00}67.31 & \cellcolor[rgb]{0.54,0.54,1.00}91.09     & \cellcolor[rgb]{0.90,0.90,1.00}9 \\
\midrule
Llama 3.3 70B & 6.86 & \cellcolor[rgb]{0.69,0.69,1.00}62.08 & \cellcolor[rgb]{0.64,0.64,1.00}71.15 & \cellcolor[rgb]{0.54,0.54,1.00}92.34  & \cellcolor[rgb]{1.00,1.00,1.00}11 \\
Llama 3.1 70B & 0.00 & \cellcolor[rgb]{0.69,0.69,1.00}61.08 & \cellcolor[rgb]{0.65,0.65,1.00}70.38 & \cellcolor[rgb]{0.59,0.59,1.00}82.59   & \cellcolor[rgb]{0.80,0.80,1.00}7 \\
    \bottomrule
\end{tabular}
\end{table}

\subsection{Tabletop Games Framework (TAG)}
\label{sec:tag}
The TAG framework is a bespoke Java research framework that supports the implementation of multiplayer tabletop board games. 
This introduces a number of new challenges:
\begin{itemize}
    \item The games are in general more complex than the simple one-player games in previous sections. 
    \item They are also inherently multiplayer. As such there is implicit opponent modeling required for good play strategies. The environment is no longer a `simple' stationary MDP, but is adversarial.
    \item The TAG framework has a number of local libraries and coding conventions; for example decks of cards are implemented via \texttt{Deck<>} or \texttt{PartialObservableDeck<>} parameterised classes. These are not likely to be present in the LLM training data, and require the LLM to generalise to unseen software architecture details. This contrasts to standard Python with common libraries of the games in earlier sections.
    \item The language used is now Java. Integration of all language models apart from Llama used \textit{langchain4j}\footnote{\url{https://docs.langchain4j.dev}}. Llama experiments used the Gemini Vertex AI interface to access the Google Llama model-as-a-service\footnote{https://console.cloud.google.com/vertex-ai/publishers/meta/model-garden}.
\end{itemize}

Algorithm~\ref{alg:framework} was applied to 12 tabletop board games (see Table~\ref{tab:TAGGames}) implemented in TAG.
These are adversarial multi-player environments with partial observability and stochasticity and varying levels of complexity. 
One player count in the supported range for each game was selected for use to give an even distribution of player counts between 2 and 4.

\begin{table}[!t]
    \centering
    \caption{TAG results by game. S.Iter is the percentage of iterations that resulted in working code and S.Trl the percentage of Trials (each consisting of 10 iterations). P is the number of players, Best Agent records the model that won the round robin tournament. SM is the number of models that produced any working code and entered an agent in the round robin tournament. BB indicates if the best agent significantly \textbf{B}eats the \textbf{B}aseline agent (OSLA or MCTS); $\approx$ means performance matches the baseline.}
    \label{tab:TAGGames}
    \begin{tabular}{lcccclr}
    \toprule
        Game & P & S.Iter & S.Trl & SM & Best Agent &  BB \\
        \midrule
        Can't Stop & 3 &  65\% & 91\% & 11 & o1-mini & Yes\\
        Colt Express & 3 & 19\% & 44\%  & 10 & Llama 3.3 70B & Yes \\
        Connect 4 & 2 & 61\% & 88\% & 11 & o1-mini & $\approx$ \\
        Diamant & 4 & 39\% & 89\%  & 11 & GPT 4o mini & Yes \\
        Dominion & 3 & 13\% & 44\% & 7 & Gemini Flash & Yes \\
        Hearts & 4 & 31\% & 65\% & 10 & GPT 4o & Yes \\
        Love Letter & 3 & 19\% & 37\% & 9 & o1-mini & $\approx$ \\
        Poker & 4 & 18\% & 50\%  & 11 & GPT 4o mini & Yes \\
        Seven Wonders & 4 & 25\% & 40\% & 7 & GPT 4o & Yes \\
        Sushi Go! & 4 & 30\% & 58\% & 10 & Gemini Flash & Yes \\
        Tic-Tac-Toe & 2 & 67\% & 87\% & 11 & No differences & Yes \\
        Virus & 2 & 29\% & 55\% & 8 & Gemini Flash &  Yes \\
        \midrule
    \end{tabular}
\end{table}

Given the additional level of complexity of these games, and different (and often dynamic) action spaces for each game, the language models were not asked to write a full policy to play the game.
Instead they were asked to write a heuristic function to estimate the value of a game state for a player. This should be close to 1.0 for a position that is a definite win, to 0.0 for a position that is a definite loss (other heuristics are possible, taking account of relative scores or ordinal positions, but this simple win/loss estimate keeps things simple).
This heuristic function was then used within a search algorithm; either one step lookahead (OSLA) or Monte Carlo Tree Search (MCTS)~\cite{coulom_efficient_2006}.

Each of these games has very different rules and implementations in TAG. 
To achieve the target of a scalable system that required no hand-writing and tuning of LLM prompts for each new game, two new TAG-specific elements were implemented to augment the process:
\begin{enumerate}
    \item Automatic extraction of the game-specific APIs. This uses Java Reflections to extract information on the methods and associated Javadoc on the game state object.
    The entry point for this is the Class name of the main game state. All public information gathering methods on this are extracted (defined as names matching on either \texttt{get*(...)} or \texttt{is*(...)}. APIs for any class dependencies on these methods, as parameters or return values, are also extracted and this recurses until the core java libraries are reached (these are excluded).
    \item Automatic rulebook digestion. This takes as input the text in the game rulebook. An approach inspired by~\cite{LLMCrafter} is used. The PDF rulebook is first broken down into chunks of 1000 or 2000 words (so any icons or pictures are excluded). The LLM is then given each chunk in turn and asked to summarise in 200 words or less the information about the game rules. 
    This final set of synopses is then fed to the LLM with a prompt to, `Summarise this information in 500 words or less.'. This provides blocks of text to include in the prompt used in the main loop of Algorithm~\ref{alg:framework} that explains the rules of the game.
\end{enumerate}

These new tools enable a scalable and game-agnostic process to be run on all games. The input for each game is the game rulebook as a PDF file, and a Java Class name for the main game state.
Additionally, the methods on the main game state were briefly reviewed for meaningful Javadoc comments, public visibility and name convention to ensure that they were picked up by the automated API process.
An example full prompt (for Sushi Go!) is included in the code repository.

The multi-player nature of these environments also necessitates a change in the evaluation criterion in Algorithm~\ref{alg:framework}. Evaluation used a tournament of 500 games between the new agent and a base agent. The base agent was either a one step lookahead (OSLA) agent (for Tic-Tac-Toe and Virus) or a vanilla MCTS agent (all other games) with a budget of 10ms and a rollout of 10 actions before the generated heuristic function is applied to estimate the value of the state. TAG uses Multiplayer-UCT with the heuristic applied for each player independently~\cite{sturtevant_analysis_2008}. This small budget enables the large number of experiments to be run in a reasonable time, but will not give the best possible players, although we go get perfect play in Tac-Tac-Toe at about 30ms with no heuristic. It remains an open question how this might change with a larger budget, but what is important here is the comparative performance of the generated heuristics.

A base opponent used the same OSLA or MCTS settings and a heuristic function of the game score normalised to [0, 1]. Tic-Tac-Toe and Connect 4 do not have scores, and the base opponents for these  rewarded a win (+1) or draw (+0.5), with 0 for a loss.
To avoid overfitting to a specific opponent all previous (working) agents to the evaluation tournament are added to later iterations within a trial. 
The evaluation score of each generated heuristic is the win rate from the most recent tournament, so this includes a broader range of opponents later in the trial. 
This performance metric means that, unlike the single-player environments, the scores from each run are not directly comparable. This modifies lines 12-17 of Algorithm~\ref{alg:framework}; \textsc{EvaluateFitness} returns the current best-performing of the agents, and this is retained as the new \emph{bestResult}. Each \emph{trial} is restarted with just the base opponent.

For each game a final tournament of 25,000 games is run between the best agents from each model, for a maximum of 11 participants if all models generated at least one heuristic that compiled and executed successfully. Points in Table~\ref{tab:TAGModels} are awarded for 1st through 10th places in this final tournament for each game. This methodology compares the results of the models directly and rewards an LLM that produces a variety of heuristics, including some that are very good at the game, over one that reliably generates valid, but near-identical code each time.

\begin{table}[]
    \centering
    \caption{TAG results by model. S.Iter and S.Trl are the same as for Table~\ref{tab:TAGModels}, plus FG is the number of games for which the LLM failed to produce any working code. Cost is the total cost of all 1200 iterations (100 per game) on the LLM. For each game in a round-robin tournament between the agents, 10 Points are given for the first place, 9 for 2nd, and so on down to 1 for 10th place. Zero points are awarded otherwise, including for LLMs that failed to produce any working code for a game. The Points column is the total of these, and Rank is the ranked order of models by points.}
    \label{tab:TAGModels}
    \begin{tabular}{lrrcrrc}
    \toprule
        Model & S.Iter & S.Trl & FG & Cost (\$) & Points & Rank \\
\midrule
Claude Sonnet  & \cellcolor[rgb]{0.92,0.92,1.00}17\% & \cellcolor[rgb]{0.59,0.59,1.00}82\% & 0 & 134.47 & \cellcolor[rgb]{0.65,0.65,1.00}61 & \cellcolor[rgb]{0.75,0.75,1.00}6  \\
Claude Haiku  & \cellcolor[rgb]{0.96,0.96,1.00}8\% & \cellcolor[rgb]{0.75,0.75,1.00}50\% & 0 & 86.71 & \cellcolor[rgb]{0.80,0.80,1.00}36 & \cellcolor[rgb]{0.90,0.90,1.00}9 \\
\midrule
Gemini Flash  & \cellcolor[rgb]{0.73,0.73,1.00}54\% & \cellcolor[rgb]{0.61,0.61,1.00}78\% & 0 & 1.60 & \cellcolor[rgb]{0.50,0.50,1.00}84 &  \cellcolor[rgb]{0.50,0.50,1.00}1st \\
Gemini Lite  & \cellcolor[rgb]{0.86,0.86,1.00}28\% & \cellcolor[rgb]{0.74,0.74,1.00}51\% & 0 & 2.93 & \cellcolor[rgb]{0.64,0.64,1.00}63 & \cellcolor[rgb]{0.70,0.70,1.00}5  \\
\midrule
Mistral Large  & \cellcolor[rgb]{0.92,0.92,1.00}16\% & \cellcolor[rgb]{0.83,0.83,1.00}34\% & 2 & 19.73 & \cellcolor[rgb]{0.86,0.86,1.00}26 &  \cellcolor[rgb]{0.95,0.95,1.00}10 \\
Mistral Small  & \cellcolor[rgb]{0.92,0.92,1.00}17\% & \cellcolor[rgb]{0.85,0.85,1.00}30\% & 7 & 3.24 & \cellcolor[rgb]{0.88,0.88,1.00}20 &  \cellcolor[rgb]{1.00,1.00,1.00}11  \\
\midrule
o1-mini  & \cellcolor[rgb]{0.71,0.71,1.00}58\% & \cellcolor[rgb]{0.62,0.62,1.00}75\% & 1 & 23.92 & \cellcolor[rgb]{0.58,0.58,1.00}69 &\cellcolor[rgb]{0.60,0.60,1.00}3rd \\
GPT 4o  & \cellcolor[rgb]{0.80,0.80,1.00}41\% & \cellcolor[rgb]{0.69,0.69,1.00}62\% & 1 & 26.01 & \cellcolor[rgb]{0.58,0.58,1.00}72 & \cellcolor[rgb]{0.55,0.55,1.00}2nd  \\
GPT 4o mini  & \cellcolor[rgb]{0.84,0.84,1.00}32\% & \cellcolor[rgb]{0.68,0.68,1.00}65\% & 1 & 4.99 & \cellcolor[rgb]{0.60,0.60,1.00}67 & \cellcolor[rgb]{0.65,0.65,1.00}4  \\
\midrule
Llama 3.3 70B  & \cellcolor[rgb]{0.69,0.69,1.00}63\% & \cellcolor[rgb]{0.59,0.59,1.00}82\% & 2 & 3.51 & \cellcolor[rgb]{0.72,0.72,1.00}48 &  \cellcolor[rgb]{0.80,0.80,1.00}7  \\
Llama 3.1 70B  & \cellcolor[rgb]{0.76,0.76,1.00}49\% & \cellcolor[rgb]{0.68,0.68,1.00}65\% & 1 & 0.00 & \cellcolor[rgb]{0.77,0.77,1.00}41 &   \cellcolor[rgb]{0.85,0.85,1.00}8  \\
\bottomrule
    \end{tabular}
\end{table}

Table~\ref{tab:TAGModels} summarises the results by language model. The Gemini 2.0 Flash model does best overall in the final tournament, generates valid code consistently and is much cheaper than the OpenAI models that also do very well. The older Mistral models do least well on these tasks on all measures; the larger Llama model is best at generating working code, but the performance of its best agents is relatively poor; the Anthropic models are much more expensive than the others, but this is not reflected in performance levels.
The reason the smaller Gemini model is more expensive (it was 25\% cheaper per token) is less working code was generated (S.Iter) so there were many more calls back to the model to fix compilation errors.

Large models do better on average than their smaller counterpart in terms of both the number of successful iterations, trials and in the quality of the best heuristics produced. 
However, in most cases the performance differences are small and this effect is smaller than the differences between the model families. GPT 4o mini is a fifth of the cost of GPT 40, but still writes the best agent in 2 games, and is only a few points behind in tournament points.

The fact that many trials fail to produce any working code over 10 iterations show the importance of re-starts, as using a single trial for each game led to more random results.



Overall results by game are shown in Table~\ref{tab:TAGGames}.
One common reason for failure of an iteration was code that compiled but then failed to execute in all edge cases due to poor error checking for division by zero (throwing a runtime error during the evaluation tournament was counted as a failure of the iteration).
There was no clear pattern of performance improvement across the 10 iterations of each trial. In some cases the later heuristics were better than the first attempts, but equally often the overall winner was the first heuristic found and later changes did not improve performance. The benefit of running more iterations for improved performance is investigated further in Section~\ref{sec:1000Minatar}.
The complexity of API to an LLM is not always the same as complexity of a game. 

Otherwise the models were often creative in their invention of undocumented API methods causing compilation to fail.
The game with the highest trial success rate, Can't Stop, is also the only game for which the base game state has no dependencies outside the core \texttt{java.lang} and \texttt{java.util} libraries, reducing the opportunities for LLMs to hallucinate about non-existent methods.

\begin{figure*}[t]
    \centering
    \includegraphics[width=0.32\linewidth]{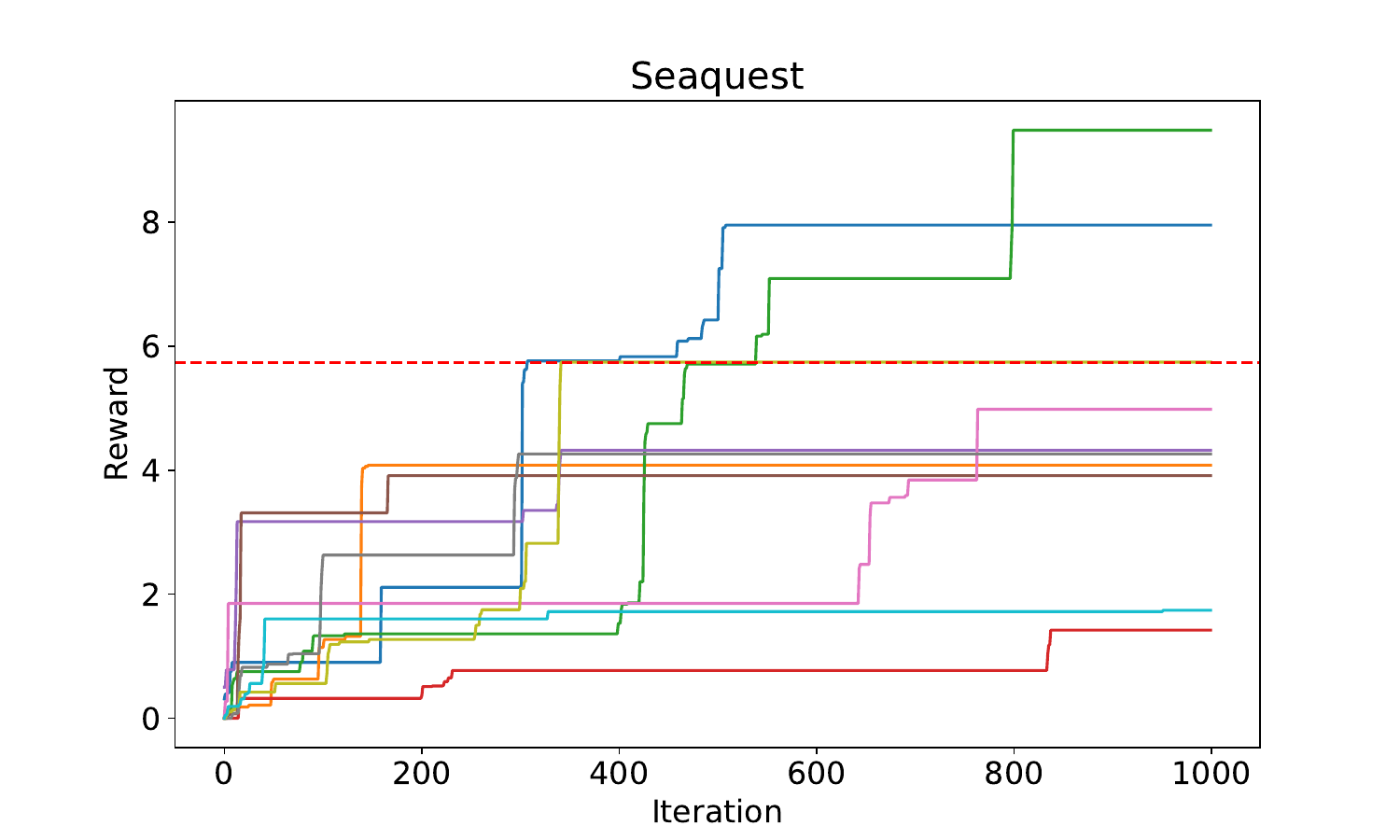}
    \includegraphics[width=0.32\linewidth]{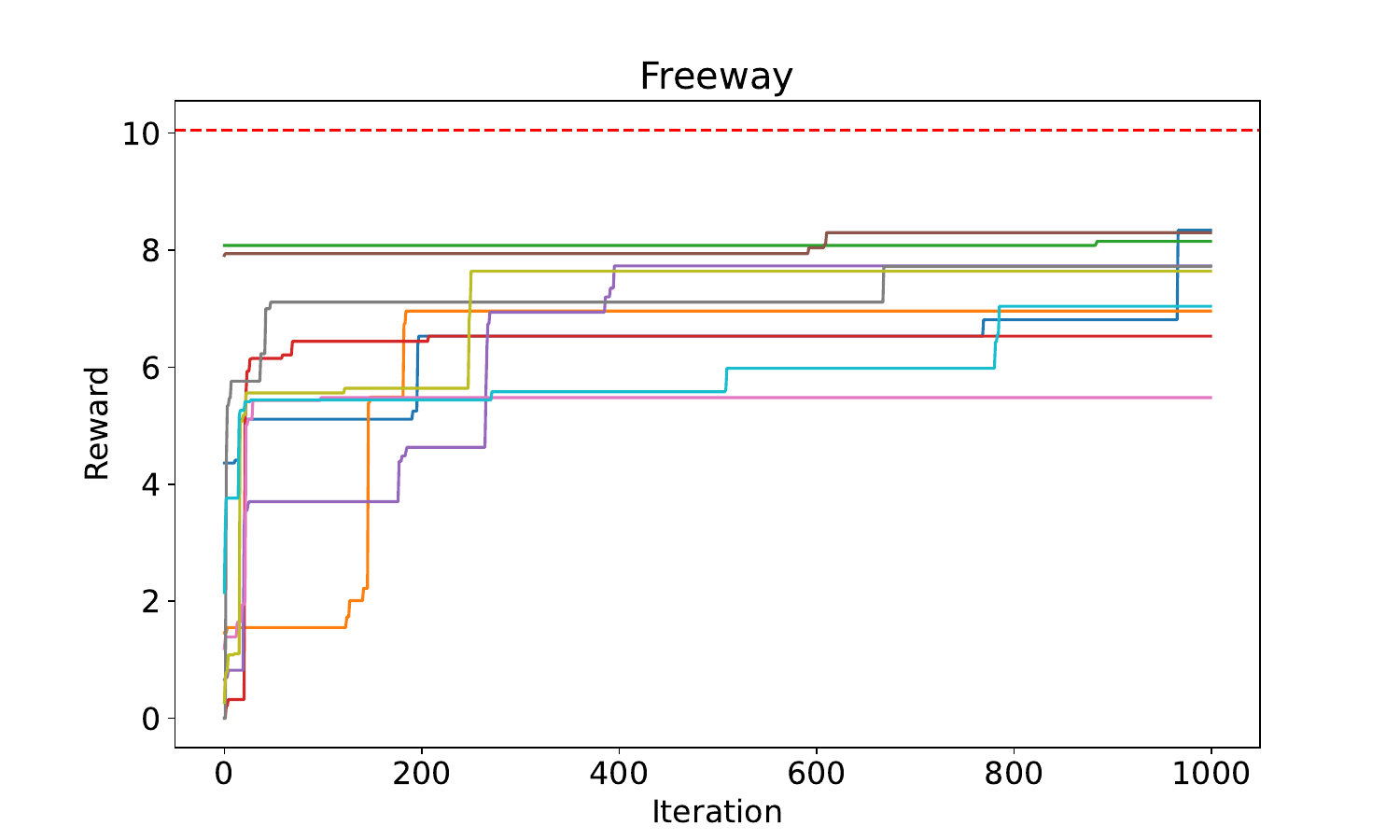}
    \includegraphics[width=0.32\linewidth]{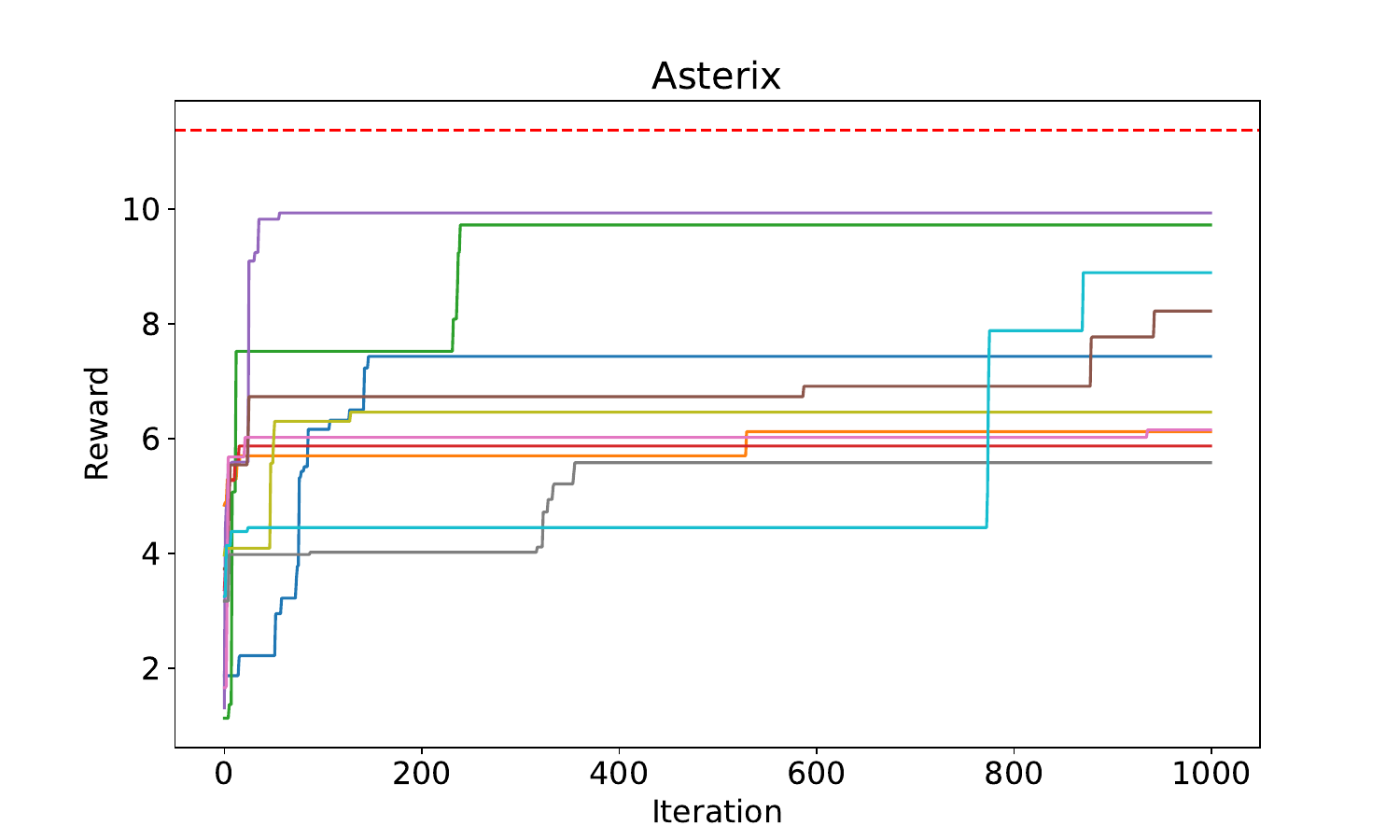}
    \includegraphics[width=0.32\linewidth]{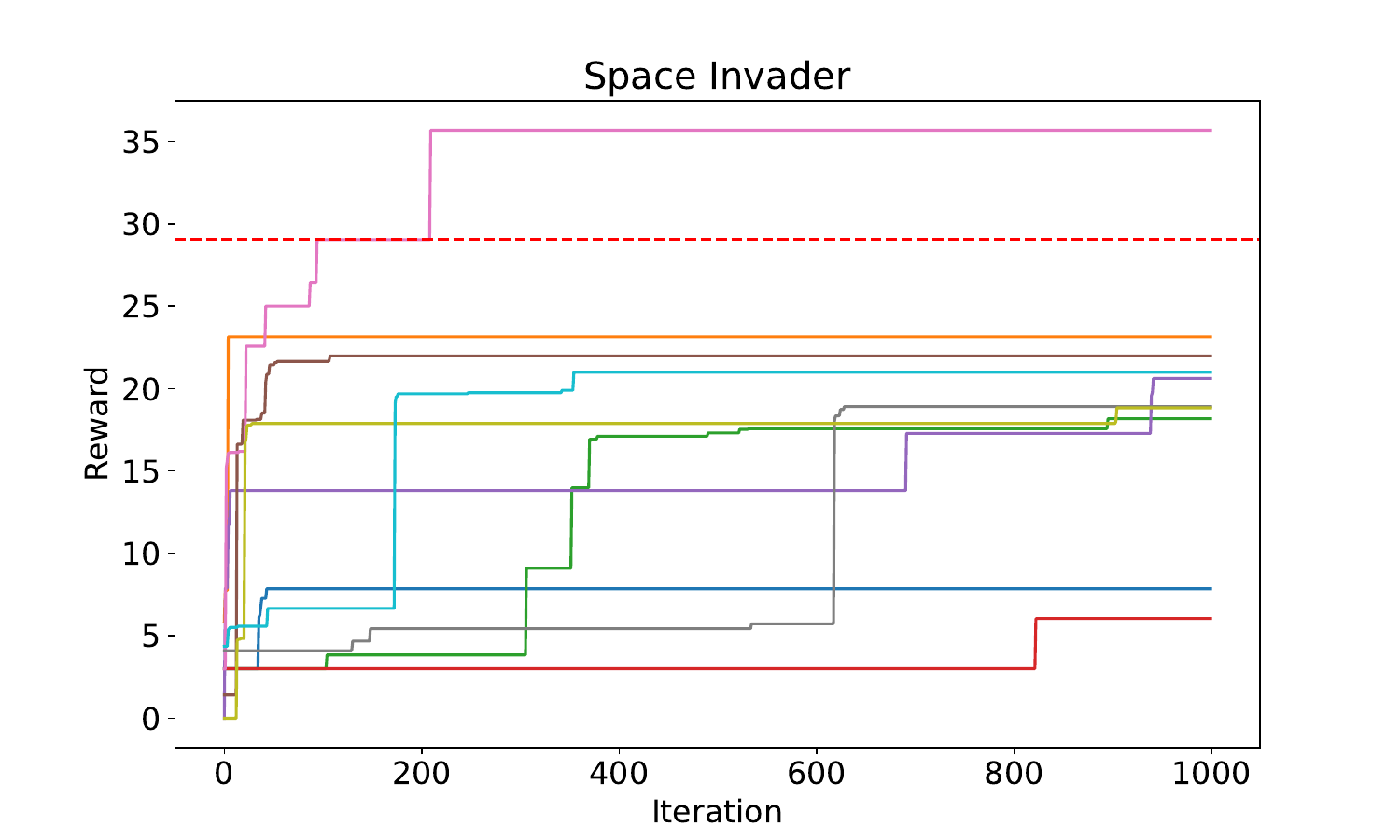}
    \includegraphics[width=0.32\linewidth]{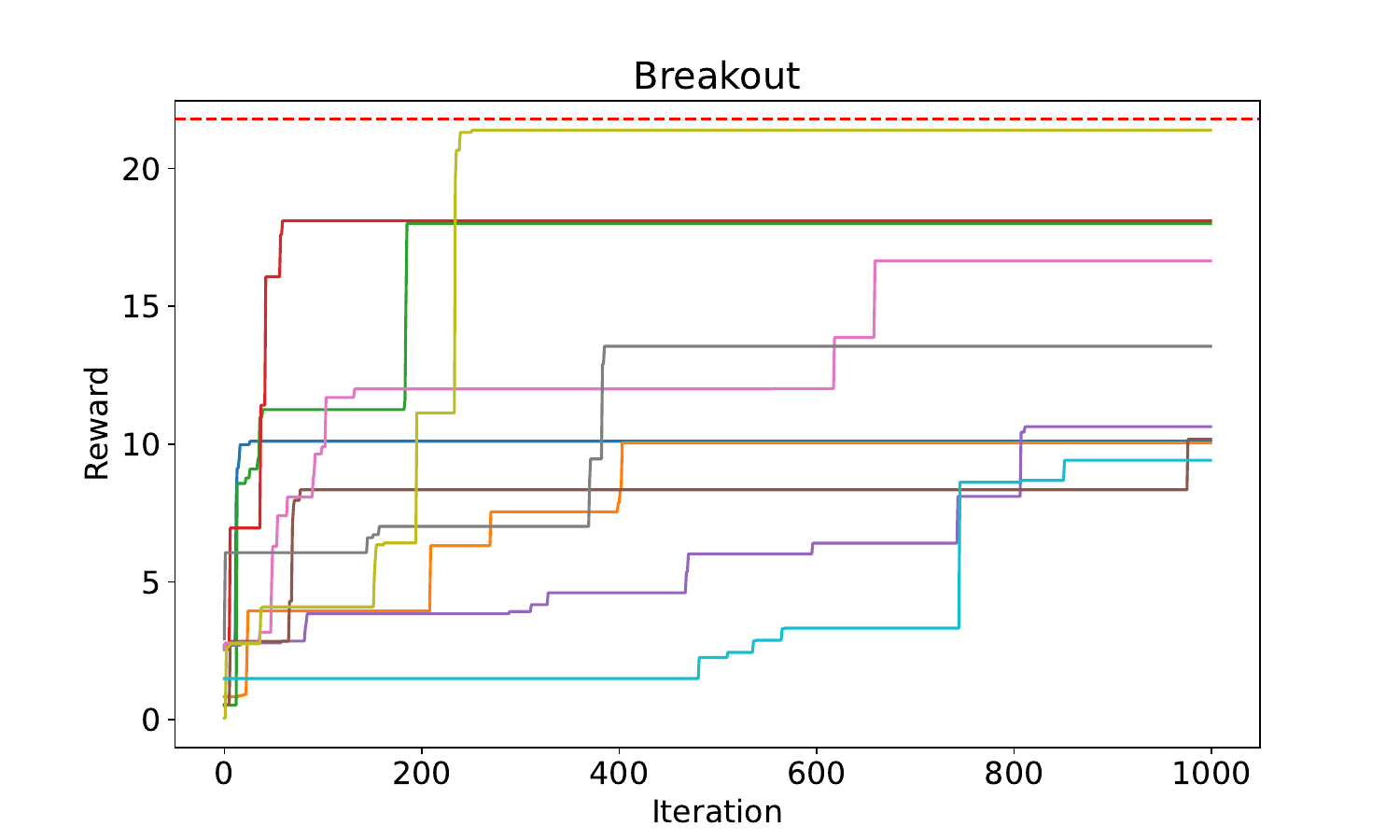}
    \caption{The reward curves for each Minatar environment for the long-running experiments. The x-axis shows the number of iterations and the y-axis the reward of the best program. The dashed red-line shows the max. reward achieved by the best LLM from Table \ref{tab:minatar-evaluation}. We use the same parameters as in \Cref{sec:exp-setup}, only with 1000 iterations instead of 10 per trial. }
    \label{fig:minatar-1000}
\end{figure*}

In contrast the card games in the set cause much more LLM code to fail to compile. All the 8 games in Table~\ref{tab:TAGGames} with \emph{S.Iter} below 35\% use decks of cards, and none of the 4 games with \emph{S.Iter} above this do. The TAG-specific infrastructure of \texttt{Deck<>} and \texttt{PartialObservableDeck<>} parameterised classes, with their own interfaces often caused problems. For example \texttt{Deck.getSize()} is used to return the number of cards, and despite this being mentioned in the prompt through recursion of the class hierarchy, many LLMs tried to call the non-existent \texttt{Deck.size()} or \texttt{Deck.length()}.
A similar problem in Sushi Go! was that the card types in the game were represented by an \texttt{enum} that had values expressed in lower case; \texttt{Maki, Sashimi, Dumpling}. Despite these values being clearly stated in the Java API section of the prompt generated code more commonly used \texttt{MAKI, SASHIMI, DUMPLING}. This is presumably because an upper case convention is more standard across Java more generally and hence in the training data of the models.


Table~\ref{tab:TAGGames}, column \textbf{BB}, shows that at least one of the LLMs wrote a heuristic that could reliably beat the baseline score heuristic in 10 of the 12 games, and was at least competitive in the other two.
Running just one trial per game per LLM (10 iterations) does not reliably get baseline performance, and restarting is needed to get a good end result from multiple independent attempts.

\section{A Thousand Iterations for Minatar}\label{sec:1000Minatar}
To evaluate the impact of the number of iterations on program search, we perform 1000 iterations for each game available in Minatar. We keep the same values for the other parameters, i.e. 10 trials for the program search and three attempts for generation/improvement or repair of non-executable programs. Figure \ref{fig:minatar-1000} shows the reward curves for each environment for the long-running experiments. In most games (with the exception of Freeway and Asterix) there is a high variance in the rewards achieved in each trial, suggesting that more restarts are better than running for more iterations. The reward curves for Freeway support this statement, since in some trials, the programs found after early iterations are almost as good as after the last iterations. 
It is also clear that the max. reward of the best trial achieves a better result for Seaquest and Space Invader and approaches the same reward for Breakout as the best LLM from \Cref{tab:minatar-evaluation}. For Freeway, the LLM is not able to improve the reward compared to the 10 iterations experiment, while for Asterix the reward is almost doubled but still not as good as the best LLM from the previous experiments. 

\section{Practical Recommendations \& Discussion}
\label{sec:practical}
\begin{table}[]
    \centering
    \caption{The ranking of the LLMs on the different domains with the overall ranking of all experiments. It is clear there are major differences between the domains.}
    \begin{tabular}{lcccccc}
    \toprule
    LLM & Minatar & Maze  & Baba & Vehicle & TAG & Overall \\
    \midrule
Claude Sonnet & \cellcolor[rgb]{0.60,0.60,1.00}3rd & \cellcolor[rgb]{0.75,0.75,1.00}6 & \cellcolor[rgb]{0.50,0.50,1.00}1st & \cellcolor[rgb]{0.65,0.65,1.00}4 & \cellcolor[rgb]{0.75,0.75,1.00}6 & \cellcolor[rgb]{0.60,0.60,1.00}3rd \\
Claude Haiku & \cellcolor[rgb]{0.80,0.80,1.00}7 & \cellcolor[rgb]{0.65,0.65,1.00}4 & \cellcolor[rgb]{0.80,0.80,1.00}7 & \cellcolor[rgb]{0.70,0.70,1.00}5 & \cellcolor[rgb]{0.90,0.90,1.00}9 & \cellcolor[rgb]{0.75,0.75,1.00}6 \\
\midrule
Gemini Flash & \cellcolor[rgb]{0.65,0.65,1.00}4 & \cellcolor[rgb]{0.80,0.80,1.00}7 & \cellcolor[rgb]{0.80,0.80,1.00}7 & \cellcolor[rgb]{0.80,0.80,1.00}7 & \cellcolor[rgb]{0.50,0.50,1.00}1st & \cellcolor[rgb]{0.65,0.65,1.00}4 \\
Gemini Lite & \cellcolor[rgb]{1.00,1.00,1.00}11 & \cellcolor[rgb]{0.70,0.70,1.00}5 & \cellcolor[rgb]{0.80,0.80,1.00}7 & \cellcolor[rgb]{0.90,0.90,1.00}9 & \cellcolor[rgb]{0.70,0.70,1.00}5 & \cellcolor[rgb]{0.95,0.95,1.00}10 \\
\midrule
Mistral Large & \cellcolor[rgb]{0.75,0.75,1.00}6 & \cellcolor[rgb]{0.95,0.95,1.00}10 & \cellcolor[rgb]{0.60,0.60,1.00}3rd & \cellcolor[rgb]{0.50,0.50,1.00}1st & \cellcolor[rgb]{0.95,0.95,1.00}10 & \cellcolor[rgb]{0.70,0.70,1.00}5 \\
Mistral Small & \cellcolor[rgb]{0.90,0.90,1.00}9 & \cellcolor[rgb]{0.95,0.95,1.00}10 & \cellcolor[rgb]{0.60,0.60,1.00}3rd & \cellcolor[rgb]{0.55,0.55,1.00}2nd & \cellcolor[rgb]{1.00,1.00,1.00}11 & \cellcolor[rgb]{0.90,0.90,1.00}9 \\
\midrule
o1 mini & \cellcolor[rgb]{0.50,0.50,1.00}1st & \cellcolor[rgb]{0.50,0.50,1.00}1st & \cellcolor[rgb]{0.55,0.55,1.00}2nd & \cellcolor[rgb]{0.75,0.75,1.00}6 & \cellcolor[rgb]{0.60,0.60,1.00}3rd & \cellcolor[rgb]{0.50,0.50,1.00}1st \\
GPT 4o & \cellcolor[rgb]{0.55,0.55,1.00}2nd & \cellcolor[rgb]{0.55,0.55,1.00}2nd & \cellcolor[rgb]{0.60,0.60,1.00}3rd & \cellcolor[rgb]{0.85,0.85,1.00}8 & \cellcolor[rgb]{0.55,0.55,1.00}2nd & \cellcolor[rgb]{0.55,0.55,1.00}2nd \\
GPT 4o mini & \cellcolor[rgb]{0.70,0.70,1.00}5 & \cellcolor[rgb]{0.55,0.55,1.00}2nd & \cellcolor[rgb]{1.00,1.00,1.00}11 & \cellcolor[rgb]{0.95,0.95,1.00}10 & \cellcolor[rgb]{0.65,0.65,1.00}4 & \cellcolor[rgb]{0.75,0.75,1.00}6 \\
\midrule
Llama 3.3 70B & \cellcolor[rgb]{0.85,0.85,1.00}8 & \cellcolor[rgb]{0.85,0.85,1.00}8 & \cellcolor[rgb]{0.80,0.80,1.00}7 & \cellcolor[rgb]{1.00,1.00,1.00}11 & \cellcolor[rgb]{0.80,0.80,1.00}7 & \cellcolor[rgb]{1.00,1.00,1.00}11 \\
Llama 3.1 70B & \cellcolor[rgb]{0.95,0.95,1.00}10 & \cellcolor[rgb]{0.90,0.90,1.00}9 & \cellcolor[rgb]{0.60,0.60,1.00}3rd & \cellcolor[rgb]{0.60,0.60,1.00}3rd & \cellcolor[rgb]{0.85,0.85,1.00}8 & \cellcolor[rgb]{0.85,0.85,1.00}8 \\
    \bottomrule
    \end{tabular}
    \label{tab:global_ranking}
\end{table}
\Cref{tab:global_ranking} shows a global ranking of the LLMs for the different domains. It is clear that there are major differences between the model families; for example, Mistral models are best for vehicle driving, while the OpenAI models are best for generating mazes and the Minatar domain. o1 mini is overall the best model, which is expected since it is the only reasoning model, and thus uses more compute than the others. For generating code in Java for TAG, Gemini 2.0 Flash is the best model followed by the GPT model family. A rather unexpected result is that Llama 3.3 is worse than Llama 3.1, although Meta advertises that it achieves a performance comparable to the Llama 3.1 405B model. In general, the open-source Llama models cannot keep up with the closed-source models of the LLM providers. However, the reasons for this are difficult to analyse as we do not know the parameter size of the paid models and there are also differences in the training pipeline and dataset. The 70B-Llama models are probably larger than the small closed-source models, but much smaller than the large models. With enough iterations and restarts, the 70B model of Llama 3.1 achieves the same performance as the large models of OpenAI and Claude, but only with 100 times more iterations. This indicates that it is in principle possible to achieve the same performance with open-source models, they just need more time to find good programs.

After discussing the game applications and the long-running experiments, there are several practical recommendations we can make for using LLMs to synthesize programs for game research. These depend primarily on the available financial resources, hardware and time.

If money is no issue, it is best to try different models and to start with the larger models as good programs are found more quickly.
With financial constraints, but local hardware to run the Llama 3.1 or 3.3 70B model, e.g. on a 48GB GPU with 4-bit quantization \cite{dettmers2023qlora}, then it is better to run local models for more iterations, as a similar reward to the Claude or OpenAI LLMs can be achieved. However, depending on the simulation environment, this may take much longer.  

If there are financial and time constraints, we recommend using \Cref{tab:global_ranking} to select the best models for a given domain.

One unexpected finding is that larger models are not always better, e.g. in the maze generation experiments, where most smaller models were  better than their larger counterpart. Therefore, smaller models are worth trying and should not be dismissed from the start. They can also be much more cost-efficient overall even if they are run for many more trials/iterations.
The results in Table~\ref{tab:TAGModels} make clear that total model costs can vary by 2 orders of magnitude with the cheaper model giving better results on a particular problem.

From the long-running experiments it is clear that more restarts are better than always running more iterations. The number of iterations depends heavily on the problem domain, and more difficult problems also need more iterations. This is visible in Figure \ref{fig:minatar-1000} for Seaquest, the most complicated Minatar domain, where the reward increases only in later iterations compared to Freeway or Asterix.

\section{Conclusion}
\label{sec:conclusion}
In this work we studied and evaluated the current possibilities of using LLMs for program search in the area of games for various applications. Previous work was mostly limited to a single problem or game without being easily transferable to other domains, as the DSL had to be adapted. 
We demonstrated that LLMs can overcome the problem of combinatorial explosion of search spaces constructed with predefined DSLs, and that LLMs are able to synthesize programmatic policies in Python for the Minatar domain, which was not possible with a custom DSL and previous methods.
Furthermore, we have shown that this framework can be easily adapted to different applications by modifying the prompts, and that it often provides reasonable results even without much customization. 
We have shown that even with the default temperature settings on these standard language models there is a very wide range of output for the same input prompt; in this respect at least the models can be quite `creative'. 
Running many independent iterations of the same task can create a varied population of outputs. This is very promising as it provides the variation required for the hill-climbing approach used here. 

We observed limitations in the quality of the generated code.  For example, in the simple 2D vehicle driving task, the generated code drove the car to the target but then failed to stop most of the time.  Much of the generated code fails to achieve any reward at all, or in the case of Java, to compile.
These limitations become more evident as the complexity of the task increases. The need to use framework-specific Java libraries in TAG leads to less than 1 in 5 attempts generating valid code. We believe limitations such as this could be overcome with more sophisticated search and better prompt engineering, but the results so far give an idea of the limitations of what can be achieved with relatively little effort.
The addition of tools to the LLM interfaces and a more agentic workflow is a promising area for this future work. For example instead of asking the LLM to generate the code in one pass, it could be asked to construct useful component functions or sub-modules with documented interfaces. In a later pass the model could then be asked to combine these sub-modules (based on feedback of performance of previous combinations).


\section{Acknowledgements}
This work stems from a working group in the Dagstuhl Seminar 24261 (Computational Creativity for Game Development, 2024), and it was supported by the EPSRC Centre for Doctoral Training in Intelligent Games \& Games Intelligence (IGGI) (EP/S022325/1).

\bibliographystyle{IEEEtran}
\bibliography{main}

\end{document}